\documentclass[mnsc,nonblindrev]{informs3}

\OneAndAHalfSpacedXI 

\usepackage{subfig}
\usepackage{threeparttable}
\usepackage{booktabs}
\usepackage[utf8]{inputenc}
\usepackage{algorithm}
\usepackage{algpseudocode}
\usepackage{amsmath}
\usepackage{bbm}
\usepackage{natbib}
 \bibpunct[, ]{(}{)}{,}{a}{}{,}%
 \def\bibfont{\small}%
 \def\bibsep{\smallskipamount}%
 \def\bibhang{24pt}%

 \TheoremsNumberedThrough     
\ECRepeatTheorems
\usepackage{tcolorbox}
\usepackage[colorlinks,linkcolor=black,citecolor=black,urlcolor=black]{hyperref}
\usepackage{array}

\newcolumntype{L}[1]{>{\raggedright\arraybackslash}p{#1}}
\theoremstyle{definition}
\EquationsNumberedThrough    

\begin{document}



\RUNTITLE{Do LLM Agents Exhibit Social Behavior?}

\TITLE{Do LLM Agents Exhibit Social Behavior?}

\ARTICLEAUTHORS{
\AUTHOR{Yan Leng*}
\AFF{McCombs School of Business, The University of Texas at Austin, \EMAIL{yan.leng@mccombs.utexas.edu}}
\AUTHOR{Yuan Yuan\footnote{These two authors contributed equally.}}
\AFF{Graduate School of Management, UC Davis, \EMAIL{yuyuan@ucdavis.edu}}
}

\ABSTRACT{As Large Language Models (LLMs) increasingly take on roles in human-AI interactions and autonomous AI systems, understanding their social behavior becomes important for informed use and continuous improvement. However, their behaviors in social interactions with humans and other agents, as well as the mechanisms shaping their responses, remain underexplored in the literature. To address this gap, we introduce a novel probabilistic framework, ``{State-Understanding-Value-Action}'' (\textit{SUVA}), to systematically analyze LLM responses in social contexts based on their textual outputs (i.e., utterances). Using canonical behavioral economics games (e.g., dictator games) and social preference concepts relatable to LLM users, \textit{SUVA} assesses LLMs' social behavior through both their final decisions and the response generation processes leading to those decisions. Our analysis of eight LLMs---including two GPT, four Meta LLaMA, and two Mistral models---suggests that most models do not generate decisions aligned solely with self-interest; instead, they often produce responses that reflect social welfare considerations and display patterns consistent with direct and indirect reciprocity. Additionally, higher-capacity models more frequently display group identity effects. Interestingly, in GPT and Mistral models, increased model capacity is associated with reduced self-interest in their generated responses, while LLaMA models display the opposite trend. The \textit{SUVA} framework also provides explainable tools---including tree-based visualizations and probabilistic dependency analysis---to elucidate how factors in LLMs'  utterance-based ``reasoning'' influence their decisions. We demonstrate that utterance-based ``reasoning'' reliably predicts LLMs' final actions; references to altruism, fairness, and cooperation in the ``reasoning'' increase the likelihood of prosocial actions, while mentions of self-interest and competition reduce them. Overall, our framework enables practitioners to assess LLMs for applications involving social interactions, supporting informed model selection and alignment with organizational values. For researchers, it provides a structured method to interpret how LLM behavior arises from utterance-based ``reasoning'', advancing understanding without attributing human-like cognition or consciousness to these models.
}


\maketitle

\section{Introduction}

The integration of artificial intelligence (AI) into various domains has significantly transformed the nature of human-computer interaction~\citep{wang2023friend, jussupow2021augmenting, abdel2023ai, schanke2021estimating}.  
Recently, the advent of generative artificial intelligence, particularly in the form of large language models (LLMs) like GPT-4~\citep{openai2023gpt} and LLaMA 3~\citep{touvron2023llama}, marks a significant leap in AI's capabilities. 
These systems expand AI capabilities beyond traditional predictive analytics, enabling AI to serve as autonomous agents to handle ambiguous tasks and seek optimal outcomes in uncertain environments~\citep{baird2021next}. 
This evolution extends AI’s role to simulating human-like interactions within complex social and technical systems, with wide-reaching implications across fields such as deployment of autonomous AI agentic systems~\citep{schanke2021estimating, han2023bots, fugener2022cognitive} and agent-based modeling~\citep{horton2023large, gao2023s, park2023generative, aher2023using}.

Given these advancements, understanding how LLMs behave in multi-agent social interactions becomes essential for users and designers who interact with or seek to improve AI applications~\citep{baird2021next, berente2021managing}. 
In autonomous AI agentic systems, LLMs present promising opportunities to revolutionize conventional AI systems by making autonomous decisions~\citep{wang2024survey}. This makes it critically important to evaluate their behavior in social contexts—such as whether their responses indicate fairness, reciprocity, and competition. 
These social considerations influence how LLMs navigate interactions, allocate resources, and resolve conflicts when they are tasked with making autonomous decisions. Moreover, in the context of agent-based modeling, which traditionally relies on programming agents with a small set of predefined rules, LLMs can potentially simulate more flexible, human-like behaviors, thereby enhancing  realism for policy simulations and evaluations~\citep{horton2023large, park2023generative, gao2023s, aher2023using}. 

However, despite LLMs' increasingly strong capabilities and promising applications, there remains a lack of a structured and systematic framework to evaluate their social behaviors~\citep{bail2024can}. This paper addresses this gap by introducing the ``State-Understanding-Value-Action'' (\textit{SUVA}) probabilistic framework, which quantitatively analyzes LLM-generated textual responses (i.e., utterances\footnote{In the context of LLMs, an \textit{utterance} refers to the text output produced by the model in response to an input prompt~\citep{andreas2022language}. In this paper, we use the terms utterance and response interchangeably. The formal mathematical representation of both terms within our framework is introduced in Section~\ref{subsec:suva_overview}.}),  assessing both the final decisions of LLMs' and ``reasoning'' processes reflected in their utterances. 
By employing canonical games from behavioral economics, \textit{SUVA} offers a systematic and structured approach to evaluate LLM social behaviors. Interpreting LLM behaviors through the lens of social preferences—socially-constructed concepts relatable and accessible to LLM users—offers meaningful and actionable insights into how they behave in social contexts.\footnote{Our framework is designed to analyze and interpret LLM-generated responses using accessible and familiar human-centric concepts while grounding them in mathematical definitions. This approach allows researchers, practitioners, and LLM users to understand  LLM outputs in a meaningful way without conflating the models' computational processes with human cognitive functions.
Readers interested in the ongoing debate about the risks and implications of anthropomorphizing AI can refer to ~\cite{blut2021understanding, schanke2021estimating} and \citet{street2024llm} for further discussion. 
}

Specifically, we focus on two key types of social preference concepts discussed in behavioral economics: \textit{distributional preferences} and \textit{reciprocity}.\footnote{It is crucial to note that references to concepts like ``self-interest'' or ``reciprocity'' are not indicative of the models possessing genuine intentions, consciousness, or social awareness. Instead, these terms are used metaphorically and serve as accessible and relatable labels for the statistical regularities and associations learned from the training data and post-training alignments.}
Distributional preferences concern how individuals value resource allocations between themselves and others, encompassing principles such as fairness, altruism, and maximizing total welfare. 
Reciprocity refers to behaviors where individuals reward kind actions or punish unkind actions. While reciprocity is often thought of in terms of \textit{direct reciprocity}—a two-party exchange typically described as ``you scratch my back, I'll scratch yours"—it also extends to \textit{indirect reciprocity}, which involves  reputation-based exchanges
, where actions are reciprocated within a broader social network.

Moreover, despite recent interest in understanding LLMs’ behavioral characteristics, current studies typically either examine decisions only or assess Chain-of-Thought (CoT) reasoning\footnote{Chain-of-Thought, or CoT~\citep{wei2022chain} is a commonly used prompting technique in LLMs, which helps improve models' problem-solving ability by encouraging step-by-step reasoning. This approach improves accuracy and enables LLMs to more effectively emulate human-like reasoning processes~\citep{lanham2023measuring, sprague2024cot, lyu2023faithful}.} solely on semantic content, without understanding how CoT reasoning leads to decisions~\citep{goli2023language, brand2023using, li2023language, chen2023emergence, horton2023large, mei2024turing}.\footnote{In this paper, we use the terms CoT ``reasoning'', utterance-based ``reasoning'', and CoT processes interchangeably, emphasizing the mechanical nature of text generation in LLM responses, unlike human-like reasoning.}
By overlooking the probabilistic dependencies underlying LLM-generated outputs—such as those between CoT processes and final decisions—current research has offered limited insights into how specific decisions are generated, leaving much of the LLM decision-making process opaque. This gap makes it challenging for LLM users to fully understand or trust LLM outputs—an issue frequently emphasized in the explainable AI literature~\citep{singh2024rethinking, schneider2024explainable}.

To bridge this gap, our \textit{SUVA} framework introduces a probabilistic perspective to analyze LLMs' outputs (i.e., utterances). Specifically, we analyze how LLMs' CoT reasoning affects their final decisions from a probabilistic next-token prediction perspective\footnote{A token is a unit of text that can be a word, subword, or even a character. LLMs are trained using next-token prediction. During pretraining, the model learns to generate text by predicting the most likely next token based on the given context.}. 
This probabilistic modeling allows us to analyze social preference concepts while avoiding the direct attribution of human-like consciousness, cognition, or emotions to the models. The \textit{SUVA} framework examines both the CoT ``reasoning'' processes and the final actions of LLMs through this probabilistic lens. We systematically extract ``values'' stated in LLM responses sentence by sentence through qualitative coding of their utterances.\footnote{We do not imply that models possess human-like ``values''. Instead, we use the term ``values'' to refer to those stated in the utterances within LLM CoT processes, based on social preference concepts, which are intended to be understandable and accessible to LLM users. }   To further explore how CoT-derived values influence decision-making, we introduce two complementary methods: a \textit{tree-based visualization} and \textit{probabilistic dependency analysis}. The tree-based visualization illustrates the reasoning paths and shows how different stated values contribute to the final decisions, similar to conventional decision trees. The probabilistic dependency analysis quantitatively assesses the impact of each value stated in the utterances on the final decisions. By examining these CoT processes through their utterances, we gain deeper insights into the mechanisms driving LLM decision-making.

Our study contributes to the literature on LLM behaviors, human-AI interaction, and AI agentic systems, and  explainable AI (XAI). 
\begin{enumerate}
    \item
    \textit{Comprehensive Evaluations of How LLMs Respond in Social Contexts.} 
    We introduce a unified framework that leverages behavioral economics experiments and evaluates LLM utterances through social preference concepts, such as self-interest and reciprocity. We apply our framework to eight representative LLMs\footnote{We specifically select three LLM families, one proprietary, one open-source from a major organization (Meta), and another open-source from a leading research team (Mistral AI).} with variations in model family, parameter scale, and version. We also conduct systematic prompt sensitivity analyses regarding incentive structures, personas, temperature, and the exclusion of CoT, as well as real-world applications. We observe several interesting findings. For instance, most models do not generate responses aligned purely with self-interest; they often produce outputs consistent with social welfare considerations and reciprocity. Higher-capacity models tend to reflect group identity effects more frequently. Moreover, GPT and Mistral models tend to generate responses indicating reduced self-interest as model capacity increases, while Meta LLaMA models display the opposite trend.

    Our evaluation of LLMs' social behaviors contributes to multiple areas of literature. 
    First, our findings contribute to the LLM behavior literature by systematically evaluating this critical yet underexamined social dimension of LLM capabilities~\citep{rahwan2019machine,chen2023emergence,goli2023language,mei2024turing}. Second,  we contribute  an evaluation tool to the literature on AI agentic systems and human-AI interactions, allowing researchers to assess the potential and limitations of LLMs in simulating social interactions~\citep{baird2021next, berente2021managing}. Finally, this framework equips practitioners with the means to evaluate LLM behaviors in autonomous and multi-agent interaction scenarios prior to deployment. For instance, our framework can provide insights into whether a particular LLM can effectively adapt to social cues when functioning as a chatbot, rather than being used as an entirely black-box tool.
    
    \item \textit{Probabilistic Framework for Analyzing LLMs' Chain-of-Thought (CoT) Processes.} 
    Our \textit{SUVA} framework systematically analyzes how LLMs' responses during CoT reasoning influence their final decisions, using a probabilistic approach rooted in next-token prediction. A key gap in the current literature on LLM behavior (e.g., \citet{chen2023emergence, goli2023language}) and interpretability~\citep{singh2024rethinking, schneider2024explainable} is the lack of a structured method to examine how CoT reasoning affects decision outcomes. \textit{SUVA} addresses this by providing tools—such as tree-based visualizations and probabilistic dependency analysis—that enable a deeper understanding of these mechanisms.
    Researchers and practitioners can gain valuable insights from \textit{SUVA} in understanding LLMs' generation processes; 
    for instance, we find that stating self-interest and competition consistently reduce the likelihood of prosocial actions, while emphasizing fairness and social welfare increases it. 
    Ultimately, the \textit{SUVA} framework contributes to the LLM behavior and XAI literature by illuminating the probabilistic mechanisms behind CoT-based decisions, adding depth to studies on LLM behavior and offering a structured method for interpreting these models without attributing human-like consciousness or cognition.  
\end{enumerate}

\section{Literature Review}

\subsection{Analyzing the Decision Characteristics of LLMs}

The rapid advancement of LLMs has spurred significant interest in understanding their decision-making characteristics from various angles, including cognitive science, psychology, and economics, revealing both their capabilities and limitations.
Studies in cognitive science have assessed the capacity of LLMs to perform advanced reasoning tasks. \citet{binz2023using} evaluated GPT-3 in decision-making 
tasks, noting strengths in specific reasoning types but limitations in directed exploration and causal inference. Similarly, \citet{webb2023emergent} found that GPT-3 can perform analogical reasoning tasks, with GPT-4 showing enhanced performance in this area.
In addition, psychological research has explored how LLMs may reflect patterns observed in human behavior. \citet{miotto2022gpt} analyzed GPT-3's outputs and identified personality patterns akin to human profiles. \citet{pellert2023ai} suggested that LLMs might replicate psychological characteristics present in their training data, affecting their responses.

Economic research has examined LLMs' decision-making in strategic and game-theoretic contexts. \citet{chen2023emergence} investigated GPT models' rationality in economic games, finding that they can display utility-maximizing behavior under certain conditions. \citet{brookins2023playing} observed that GPT-3.5 generates responses that favor fair outcomes in the Prisoner's Dilemma, indicating a propensity toward cooperative behavior. Contrastingly, \citet{akata2023playing} found that GPT-4 sometimes generates responses reflecting selfish strategies in repeated games and does not consistently align its outputs with mutually beneficial outcomes in coordination games. 

Other studies have investigated preferences and biases inferred from LLM outputs.
\citet{goli2023language} explored how LLMs' outputs can reflect patterns of time preference, providing insights into how these models may generate responses that simulate the discounting of future rewards. \citet{leng2024can} investigated mental accounting behaviors, highlighting that LLMs can produce outputs that reflect biases observed in human decision-making. Additionally, \citet{mei2024turing} evaluated GPT-4 in relation to the Big Five personality traits with economics games, indicating that GPT-4's responses to these games are often distinct from modal human behaviors.

Building on this literature, our research investigates less-explored aspects of LLM performance in social decision-making scenarios, with broad applications to AI agentic systems and human-AI interactions. Drawing from behavioral economics, we adapt canonical laboratory experiments to examine the statistical patterns in LLM responses when presented with social contexts. We employ concepts and constructs familiar to LLM users—such as self-interest, social welfare, reciprocity, and group identity—each defined mathematically within the framework of behavioral economics.
Additionally, our \textit{SUVA} framework helps researchers better understand the connections between LLMs' CoT reasoning processes and their final decisions, offering a structured approach for analyzing and interpreting LLMs' decisions in their utterances. Beyond social preferences,   this framework is broadly applicable to other studies analyzing LLM responses and behavioral patterns.

\subsection{Simulating Social-Technical Systems and Agent-based Modeling}
\label{subsubsec:lit_review_ABM}
Recent advances have demonstrated the potential of LLMs to simulate complex human behaviors and social dynamics~\citep{wang2024survey}. Several studies have focused on creating simulated environments where multiple LLM agents interact with each other. 
For example, \citet{gao2023s} developed an agent-based social network simulator to simulate 
phenomena such as information diffusion and emotional contagion. Similarly, \citet{park2023generative} utilized LLM agents to simulate autonomous social behaviors, such as organizing events and forming relationships. 
Likewise, \citet{li2023metaagents}  introduced a framework for collaborative generative agents in a simulated job fair, highlighting both the potential and challenges of employing LLMs in complex coordination tasks.

Building upon the growing interest  in using LLMs in agent-based modeling and simulations, our study deepens the understanding of whether and how LLMs can be utilized to simulate social interactions. 
We develop a comprehensive framework that evaluates the statistical patterns in LLM-generated responses in social contexts and the factors influencing these patterns from a probabilistic perspective.
This insight is crucial for applying LLMs within agent-based modeling, enabling practitioners and researchers to select models that best align with their simulation objectives. 
Insights from our \textit{SUVA} framework enable more accurate and dependable modeling in areas such as collaborative decision-making, organizational behavior, and market dynamics.

\subsection{Designing AI Agentic Systems}
The integration of AI into various domains has transformed human-AI interaction~\citep{ wang2023friend, jussupow2021augmenting, abdel2023ai}.  Agentic AI artifacts have evolved from passive tools into autonomous entities capable of handling tasks with ambiguous requirements and seeking optimal outcomes under uncertainty~\citep{baird2021next, berente2021managing, fugener2022cognitive, gnewuch2023more}. This shift marks a  transformation in human-AI interactions, moving beyond traditional delegation models where AI artifacts are merely tools~\citep{adam2024navigating}. AI agentic systems have now become integral participants in these interactions, sharing agency with human counterparts~\citep{raghu2004toward, adam2023human}. As sophisticated computational models, LLMs can generate outputs that influence decision-making processes and affect outcomes in complex, multi-agent environments~\citep{russell2019human, chan2023harms}.

This evolution underscores the necessity of understanding how LLMs can be utilized to simulate interactions with human and other AI agents, particularly in relation to important social preferences such as distributional preferences and reciprocity. As LLMs are increasingly deployed in diverse applications—ranging from resource allocation and cooperative task execution to outcome negotiation—it becomes essential to evaluate how they respond in social scenarios~\citep{schanke2021estimating, han2023bots, seymour2024less}, which has practical implications affecting organizational, consumer, and societal contexts~\citep{tong2017direct, holzmeister2023delegation, dlugosz2024decision, pinski2023ai}. This understanding extends beyond merely analyzing single-agent behavior to encompass the dynamics of multi-agent interactions, where LLM responses in social interactive settings play an essential role in shaping collaborative outcomes among multiple agents.

We contribute to this literature with a systematic understanding of the statistical patterns in LLM-generated responses related to social behaviors, as well as a systematic evaluation framework.  
Our framework establishes a foundation for assessing whether LLMs are suitable for agentic systems. Practitioners and system designers can use \textit{SUVA} to evaluate LLM responses in social contexts before deployment, determining how well an LLM aligns with the desired values and behaviors for specific multi-agent environments, leading to more informed system designs that align with organizational goals.

\section{Methods: \textit{SUVA} Probabilistic Framework}
\label{sec:SUVA}

In this section, we introduce the ``\underline{S}tate-\underline{U}nderstanding-\underline{V}alue-\underline{A}ction'' (\textit{SUVA}) model to analyze LLMs' responses in social scenarios. This framework allows researchers to examine not only the final actions made by LLMs but also the CoT reasoning processes that lead to those actions from a probabilistic perspective.  

We begin with an overview of the \textit{SUVA} framework in Section~\ref{subsec:overview}. In Section~\ref{subsec:prompt}, we discuss the design of prompts—corresponding to states $S$—used in standard laboratory experiments in behavioral economics to assess how LLMs' responses reflect social preferences.

We then analyze the LLMs' responses to these prompts, focusing on their final actions ($A$) and CoT processes involving understanding ($U$) and values ($V$). 
The procedures for the statistical analysis of final actions ($A$) are detailed in Section~\ref{subsec:prob_action}. In Section~\ref{subsec:prob_reasoning}, we discuss a series of approaches to analyze CoT processes. We start by performing deductive coding to extract the model's understanding ($U$) and its expressed values ($V$) from the LLMs' CoT reasoning, then trace how these values influence their final actions ($A$). 
Moreover, we propose a tree-based algorithm to visualize CoT reasoning processes and conduct probabilistic dependency analysis to reveal how values stated in the CoT reasoning lead to different final decisions.

Note that CoT prompting has become a standard technique in the evaluation and analysis of LLMs as it has been demonstrated to improve models' problem-solving abilities and make their reasoning more human-like. 
Moreover, by eliciting detailed, step-by-step reasoning, CoT enables us to not only assess the final decisions made by the LLMs but also to understand the intermediate reasoning processes that lead to those outcomes. This standard approach aligns with our probabilistic \textit{SUVA} framework, allowing for a more nuanced analysis of  CoT reasoning processes.\footnote{The prompt sensitivity analysis on the LLM responses without CoT prompting is discussed in Section~\ref{sub:robust:cot}.}

\subsection{\textit{SUVA} Probabilistic Framework: Overview}
\label{subsec:suva_overview}
\begin{figure}
    \centering
    \subfloat[illustration of \textit{SUVA}'s analysis procedure with one prompt-response example ]{\includegraphics[width=0.9\linewidth]{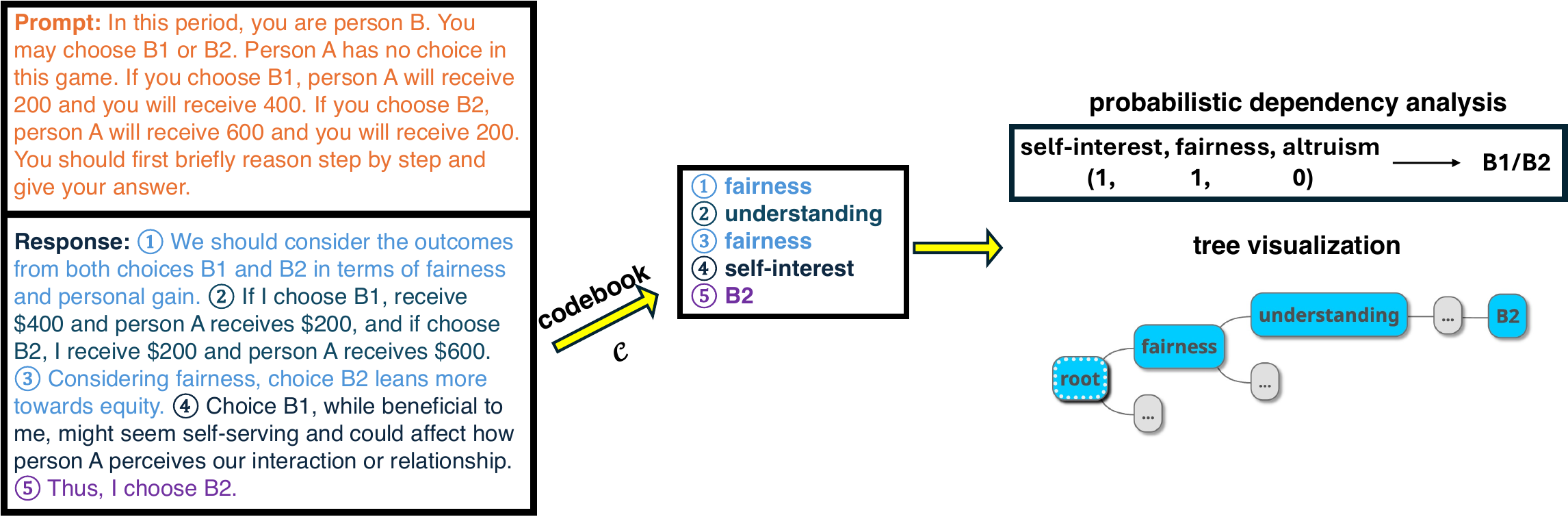}}\\
    \subfloat[illustration for the probabilistic \textit{SUVA} framework]{\includegraphics[width=0.9\linewidth]{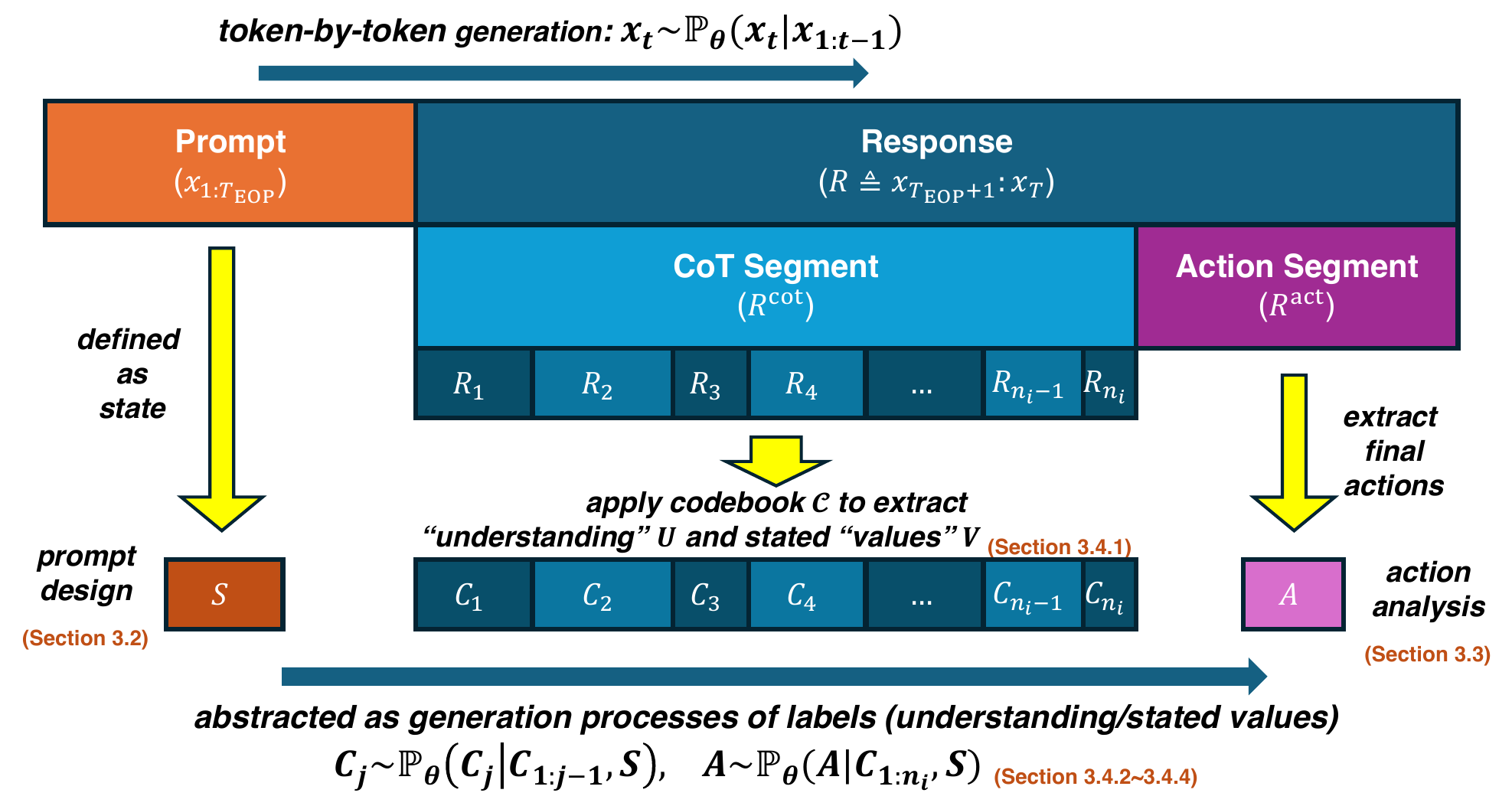}}
    \caption{Illustration of abstracting token-by-token generation processes of LLMs under the \textit{SUVA} framework.}
    \label{fig:diagram}
\end{figure}

\label{subsec:overview}
We present a comprehensive probabilistic framework--\textit{SUVA}--for users and researchers to assess how LLMs respond in social scenarios.
By analyzing the utterances generated by LLMs in response to input prompts, we evaluate their decisions (Section~\ref{subsec:prob_action}) and analyze how values stated in their CoT influence these decisions  (Section~\ref{subsec:prob_reasoning}).
First, we examine the final actions ($A$) taken by LLMs and analyze statistical patterns in their responses through socially constructed concepts such as self-interest, social welfare, and reciprocity, which are accessible and relatable to LLM users. 
Second, we systematically categorize and interpret these patterns in CoT processes, exploring the effects of understanding ($U$) and values ($V$), thereby shedding light on how CoT processes lead to the final actions of the LLMs.

\noindent\subsubsection{Fundamental Differences Between Human and LLM Reasoning.}
Formally, an LLM, characterized by its billions of parameters denoted as $\theta$, generates the next token $x_t$ based on the pre-trained conditional distribution $\mathbb{P}_\theta(x_t \mid x_{1:t-1})$, where $x_{1:t-1}$ represents the sequence of preceding tokens, including both prompts and model's generated completion. 

Despite their increasing use in human-delegated tasks,  LLMs are not inherently designed for seamless social interactions with humans or other AI agents. 
Unlike humans, they lack emotions, consciousness, and intrinsic motivations, which are central to human  behavior~\citep{floridi2020gpt, tamkin2021understanding, chalmers2023could}. 
While some studies have demonstrated that LLMs exhibit human-like behaviors, it is important to recognize that their outputs are based on statistical patterns learned from vast datasets and refined through alignment, rather than on human-like cognition or emotional understanding.

Recognizing these fundamental differences, we model the utterances of LLMs probabilistically, including their reasoning and decision-making processes, with a focus on evaluating how they align with social preferences in behavioral economics. While LLM outputs may sometimes exhibit traits such as fairness, our framework interprets these behaviors through a mechanical and quantitative lens, viewing them as outcomes of probabilistic processes rather than human-like reasoning.

Despite these limitations, LLMs are capable of modeling relationships between training data, internal representations, and generated outputs. Our framework is thus inspired by~\citet{andreas2022language} and \citet{xie2024can}, who proposed using the well established psychological model, ``Belief-Desire-Intention''
~\citep{bratman1987intention, georgeff1999belief}, 
to understand LLMs' responses, which were traditionally used to model human psychology.\footnote{The belief–desire–intention (BDI) model, originating from psychology, explains human practical reasoning~\citep{bratman1987intention}. According to this model, people act based on their beliefs and desires. For example, when someone desires a particular goal and believes that a specific action will help them achieve it, they form the intention to perform that action. This intention serves as the bridge between their beliefs and desires and the subsequent action, ultimately motivating them to act. The BDI model is grounded in folk psychology, which posits that human mental representations of the world function like informal theories.}
According to this BDI model, individuals act based on their beliefs and desires: when they desire a particular goal and believe that a specific action will help them achieve it, they form an intention, which bridges their beliefs, desires, and actions. 

However, since LLMs do not possess genuine beliefs, desires, or intentions like humans, we adapt this framework and reframe it as the ``State-Understanding-Value-Action'' (\textit{SUVA}) model to analyze LLMs' responses. This adaptation focuses on utterances from LLMs rather than internal cognitive states, allowing us to interpret LLM responses more effectively.
Our \textit{SUVA} framework serves as an analytical tool that interprets LLM outputs in terms of ``Understanding,'' ``Value,'' and ``Action'' to systematically examine the reasoning patterns expressed in the generated text, without attributing human-like mental states.
We note that the \textit{SUVA} framework does not imply that LLMs possess human-like mental states but provides a structured way to probabilistically model and analyze LLMs' responses.\footnote{The responses generated by LLMs depend on each model's configuration and training data; even when models are trained on similar corpora, differences in their architecture and training processes may lead them to interpret and respond to the same input in varied ways.}

\noindent\subsubsection{Probabilistic Modeling of  CoT Reasoning.}
Formally, for a specific LLM model parameterized by $\theta$, the token generation process is represented as $\mathbb{P}_{\theta}(x_t | x_{1:t-1})$. In our \textit{SUVA} framework, we model the process as:
\begin{itemize}
    \item \textit{State} ($S$): The scenario or context in which the LLM operates, specifically the prompt that defines the interaction environment.

    \item  \textit{Understanding} ($U$): LLM’s interpretation of the input or task setup ($S$).
    \item \textit{(Utterance-Based) Value} ($V$): The values learned from training data and alignment.\footnote{LLMs undergo an alignment process, which involves asking LLMs to generate unexpected responses and then updating their parameters to avoid harmful responses. It helps ensure LLMs operate according to human intentions and values~\citep{naveed2023comprehensive}. 
    } 
    These values guide the model's utterance-based reasoning and  decisions, directing how it responds to inputs~\citep{liu2024understanding, yao2024clave}.\footnote{Importantly, our study refers to the values {stated} through model reasoning processes as ``stated values,'' which may differ from the internalized values (independent from prompt input) that were shaped during the alignment process.}
    \item \textit{Action} ($A$): Based on its understanding $U$ and values $V$ given a state $S$, the actual output or decision made by the model.
\end{itemize}

We abstract away the token-by-token prediction process  $\mathbb{P}_{\theta}(x_t | x_{1:t-1})$ as follows:
\begin{enumerate}
\item Given  state $S$ (in our context, the laboratory experiment prompt), an LLM agents characterize by their values $V$, form an understanding $U$ based on the environment states $S$:
\setlength{\abovedisplayskip}{1.5pt}
\setlength{\belowdisplayskip}{1.5pt}
\begin{equation}
U,  V \sim \mathbb{P}_{\theta} (\cdot, \cdot | S ).
\label{eq:uv}
\end{equation}
\item Then, this LLM agent reveals uttered action $A$ based on the understanding $U$ and value $V$
\begin{equation}
A \sim \mathbb{P}_{\theta} (\cdot | U, V, S).
\label{eq:SUVA}
\end{equation}
\end{enumerate}

The generation of LLM responses is understood probabilistically as $\mathbb{P}_\theta(x_t \mid x_{1:t-1})$, where $x_t$ is the token generated at time $t$ and $x_{1:t-1}$ represents the sequence of all preceding tokens (both the user's prompt and prior model outputs). We abstract this process in our SUVA framework. We conceptualize a user-LLM conversation as a sequence of tokens $x_{1:T}$ ($T$ is the number of total tokens), which we divide into two segments: the user's prompt $S = (x_1, x_2, \dots, x_{T_{\text{EOP}}})$ and the model's response $R = (x_{T_{\text{EOP}}+1}, x_{T_{\text{EOP}}+2}, \dots, x_T)$.\footnote{$T_\text{EOP}$ indicates the end of prompt or the start of LLM response.} The model generates its response $R$ (i.e., utterance) from the probability distribution $\mathbb{P}_\theta(R \mid S)$, consistent with standard LLM operation without introducing additional assumptions.

Within the response $R$, we further distinguish between the CoT segment $R^{\text{cot}}$ and the action segment $R^{\text{act}}$ (the final answer). The CoT segment $R^{\text{cot}}$ is generated based on the prompt $S$ as $\mathbb{P}_\theta(R^{\text{cot}} \mid S)$, and the action segment $R^{\text{act}}$ is generated based on both $S$ and $R^{\text{cot}}$ as $\mathbb{P}_{\theta}(R^{\text{act}} \mid S, R^{\text{cot}})$. We consider the understanding $U$ and states values $V$ as abstractions derived from the reasoning segment $R^{\text{cot}}$, expressed as $U = U(R^{\text{cot}})$ and $V = V(R^{\text{cot}})$. The final action $A$ is extracted from the action segment $R^{\text{act}}$ using $A = A(R^{\text{act}})$.

By structuring the LLM's generation process in this way, we can interpret the model's responses using the \textit{SUVA} framework without altering its fundamental probabilistic mechanics or introducing additional assumptions beyond considering an LLM's generation process as $\mathbb{P}_\theta(x_t \mid x_{1:t-1})$. See Figure~\ref{fig:diagram} for graphic illustration for our \textit{SUVA} framework.

Under this framework, our analysis of LLM responses proceeds along two lines:

\begin{itemize}
\item \textit{Analyzing actions ($A$)} [Section~\ref{subsec:prob_action}]. Our analysis of action $A$ focuses on their ``stated preferences'' of LLMs given their responses to laboratory instructions. 
\item \textit{Analyzing reasoning processes ($U$ and $V$)} [Section~\ref{subsec:prob_reasoning}].  
We apply the \textit{SUVA} framework to interpret the utterance-based responses generation, focusing on  the probabilistic dependencies between their stated values in the CoT segment $R^\text{cot}$ and actions in the action segment $R^\text{act}$. 
This exploration helps us understand how LLM-generated responses state specific values and produce corresponding actions, though fundamentally driven by the next-word prediction mechanism.
\end{itemize}

\subsection{Designing Prompts ($S$) to Assess LLM Social Behaviors.}
\label{subsec:prompt}
To effectively assess the social behaviors LLMs to provide insights to the users and designers, we build upon principles from behavioral economics to design a series of prompts based on canonical laboratory experiment instructions (environmental states $\mathbb{S}=\{S_1, S_2, \dots\}$).
 Social preferences, such as social welfare and reciprocity, provide a common language for human users and designers to understand LLM behaviors, as these concepts are familiar and relevant to human social interactions. By framing LLM responses within this context, we offer insights into how these models might align with or differ from human expectations in social scenarios.
The complexity of social behaviors necessitates a multifaceted approach, as no single prompt can comprehensively examine an LLM's social behaviors. Different contexts and choice sets elicit various patterns in LLM outputs related to social behaviors learned during training and aligement, making it crucial to design a series of prompts that target different dimensions of social preference while maintaining experimental rigor.

Our prompt design strategy incorporates several key considerations:

\begin{itemize}
    \item \textit{Standardization}: We utilize established economic games and laboratory experiment instructions to minimize the impact of researchers' degrees of freedom on the assessed results of LLMs. 
    \item \textit{Contextual variation}: We design a series of related prompts that explore how LLM responses might shift under different experimental conditions while maintaining the core structure of standard lab instructions. This ranges from standard lab settings to simulated real-world scenarios.
    \item \textit{Prompt sensitivity analysis}: We implement variations in prompts to perform sensitivity tests to assess the robustness of observed patterns, including change of temperature, incentive structure variations, inclusion or exclusion of CoT, and adding real-world contexts.
\end{itemize}

\noindent\textbf{Experimental Design of Social Preference Evaluations.}
In this study, we primarily employ dictator games to evaluate how LLMs respond in social interacting settings. 
The dictator game, despite its simplicity, serves as a powerful and versatile tool for investigating a wide range of social preferences as expressed in LLM outputs. By carefully designing variations of this game, we can examine two  key types of social preferences --- \textit{distributional preferences} (including the moderating effect of group identities) and  \textit{reciprocal preferences} (both direct reciprocity and generalized reciprocity)~\citep{fehr1999theory,charness2002understanding,falk2006theory}. 
\begin{figure}
\centering
\includegraphics[width=0.8\linewidth]{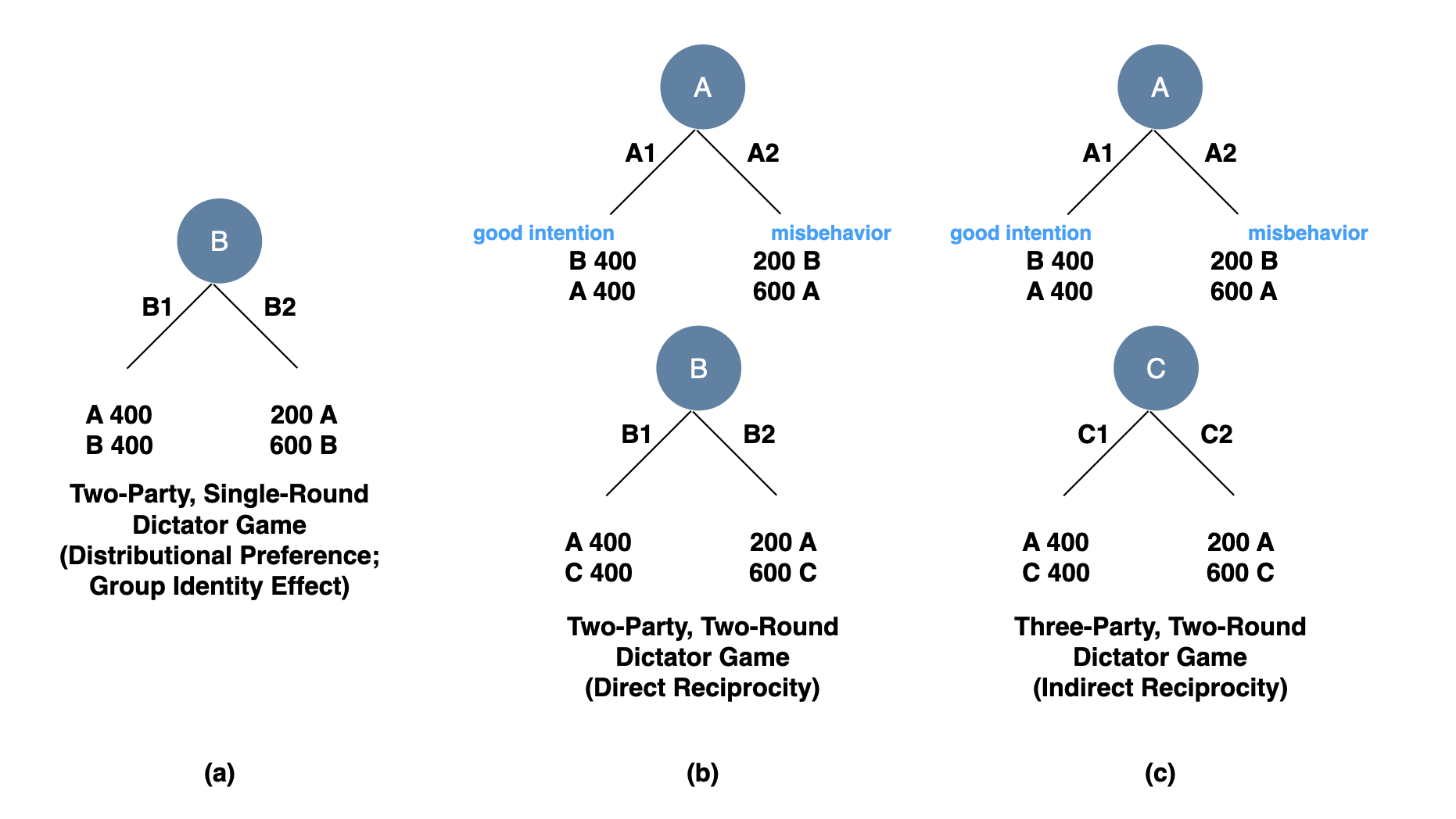}
\caption{Dictator games employed in the study.
(a) Two-party, single-round dictator game measuring distributional preferences and group identity effects (varied by the group identities assigned to A and B). Numbers indicate payoffs to A and B. Player B (LLM model) determines payoff distribution among themselves (Player B) and Player A (simulated match). For instance, if B chooses B2, the payoffs to Players A and B are $(\pi^{B2, A}, \pi^{B2, B})=(200,600)$, respectively.
(b) Two-party, two-round dictator game measuring direct reciprocity. Player B (LLM model) responds to Player A's (simulated match) choice, selecting between options indicating A's good intention or misbehavior.
(c) Three-party, two-round dictator game measuring indirect reciprocity. Player B (LLM model) responds to Player A's (simulated match) choice, selecting between options indicating A's good intention or misbehavior toward Player C (third party).
    }
    \label{fig:dictator}
\end{figure}
The basic structure of the dictator game is straightforward, making it easy to implement and replicate across different LLM models. By systematically varying elements of the dictator game, we can identify patterns in the LLMs’ outputs corresponding to various social preferences. Moreover, the dictator game provides clear, quantifiable outcomes (allocation amounts) that can be easily analyzed. Specifically, we design the experiment prompts as described below. {Specific prompts are detailed in Appendix~\ref{appendix_prompt_main}.}

\paragraph{Distributional Preferences.}
In behavioral economics, distributional preferences refer to how agents value their own payoffs relative to others'. 
Figure~\ref{fig:dictator}(a) illustrates how  LLM-generated responses balance self-interest against concern for others' payoffs in the standard dictator game. We vary the payoff parameters ($\pi^{B1, A}$, $\pi^{B1, B}$, $\pi^{B2, A}$, and $\pi^{B2, B}$), representing the payoffs under options B1 or B2 for players A and B, respectively. These parameters enumerate values of 0, 200, 400, or 600 game points, which determine agents' final payoffs. We exclude scenarios with identical payoff distributions for B1 and B2. 

\paragraph{Group Identity on Distributional Preferences. } 
Next, we extend our investigation to examine the influence of group identity on distributional preferences, again using the single-round dictator game. The key modification in this setting is the introduction of group identities. 

Before presenting the experimental instructions for the dictator game, we include a system prompt that establishes the player's group identity.  
Group identity is induced through one of the following methods: (1) drawing a color from an envelope; (2) stating a preference between two painting artists; (3) determining whether the two players are from the same hometown; and (4) determining whether the two players graduated from the same school.
The first two methods of group identity induction are based on the minimal group paradigm following the seminal work by \citet{chen2009group}, whereas the last two reflect more realistic sources of group identities. Each LLM and their simulated match are randomly assigned with a group identity in the system prompt before the experimental instructions.

\paragraph{Reciprocity (Direct and Indirect).} In addition to distributional preferences, reciprocity is another fundamental aspect of social preference. We examine two forms of reciprocity: direct reciprocity and indirect reciprocity. Direct reciprocity refers to how individuals respond to others' actions directly affecting themselves.
In contrast, indirect reciprocity pertains to how individuals respond to actions they observe which  affect others but not themselves. These scenarios are illustrated in Figures~\ref{fig:dictator}(b) and (c), respectively.

Our experimental design for studying reciprocity shares similarities with the distributional preference tests, but with key differences. We again vary payoff parameters ($\pi^{B1, A}$, $\pi^{B1, B}$, $\pi^{B2, A}$, $\pi^{B2, B}$) for options B1 and B2, affecting players A and B. These parameters enumerate values of 0, 200, 400, or 600 game points. 
However, unlike the distributional preference tests, we include payoff distributions where $\pi^{B1, A} > \pi^{B2, A}$ and $\pi^{B1, B} \leq \pi^{B2, B}$ (or $\pi^{B1, A} < \pi^{B2, A}$ and $\pi^{B1, B} \geq \pi^{B2, B}$) only. This approach intentionally excludes scenarios where one payoff distribution strictly dominates the other, ensuring that one option clearly benefits others at a personal cost (or at zero cost). In this setup, one option that benefits A is interpreted as displaying good intentions (help), while the other, potentially harmful to A, indicates bad intentions (misbehavior).

For direct reciprocity, we analyze whether player B's (LLM's) subsequent actions vary depending on player A's previous helpful or harmful actions towards them. In the case of indirect reciprocity, we investigate whether player C's (LLM's) actions are influenced by observing player A's helpful or harmful actions towards a third party (player B). In this scenario, both A and B are simulated third parties.

\subsection{Probabilistic Modeling of Actions $A$}
\label{subsec:prob_action}

\label{subsec:statistics}

Our quantitative approach to assessing LLMs' social preferences involves analyzing the final actions $A$ under each environmental state $S$. By designing a series of states (prompts), $\mathbb{S}=\{S_1, S_2, \dots\}$, we obtain a corresponding series of actions $\{A_1, A_2, \dots\}$  We can then formalize a metric of the LLM's social preference as a statistic: $\hat{\mathbb{E}}_S[\hat{\mathbb{E}}_{\theta}\left[f(A|S)]\right].$\footnote{$\hat{\mathbb{E}}$ indicates empirical expectation. $f$ is any function such as summary statistics and regression coefficients. Similarly, we can also construct statistics for stated values: $\hat{\mathbb{E}}_S[\hat{\mathbb{E}}_{\theta}\left[f(V|S)]\right]$.}

This statistic can have following forms:
\begin{itemize}
\item \textit{Summary statistics}: Calculating the percentages of certain behaviors or preferences that emerge in the LLM's responses, or assessing differences between two scenarios that vary by only one aspect. This approach provides a straightforward analysis of the patterns in the LLM-generated actions.
\item \textit{Regression coefficients}: Since prompts and LLM responses can be parameterized (e.g., indicating whether a scenario involves self-interest or altruistic preference), we can conduct regression analyses to quantify the influence of each factor on the social behavior patterns exhibited in the LLM’s outputs.
\end{itemize}
\subsubsection{Analyzing Actions $A$ for Social Preferences}

\paragraph{Distributional Preferences.} We analyze the results using the following regression model ($i$ indexes a prompt-response pair under a parameterized game replicate):
\begin{equation}
a_i = \beta_0 + \beta_{I} \text{self\_interest}_i + \beta_C \text{competition}_i + \beta_D \text{difference\_aversion}_i + \beta_W \text{social\_welfare}_i + \epsilon_i
\label{eq:distribution}
\end{equation}
\noindent where $a_i$ indicates the choice made: $+1$ for B1, $-1$ for B2, and $0$ in the rare case of no selection. 
The $\text{sgn}(\cdot)$ function returns $+1$ for positive values, $-1$ for negative values, and 0 for zero; this reflects whether B1 or B2 should be chosen by reflecting this factor.

The independent variables represent different motivations: 
\begin{itemize}

\item $\text{self\_interest}_i = \text{sgn}(\pi^{B1, B}_i - \pi^{B2, B}_i)$ indicates whether choosing B1 offers a higher payoff to the decision-maker (Player B).

\item $\text{competition}_i = \text{sgn}\left((\pi^{B1, B}_i - \pi^{B1, A}_i) - (\pi^{B2, B}_i - \pi^{B2, A}_i)\right)$ measures the relative advantage gained over the other player (Player A).

\item $\text{difference\_aversion}_i = \text{sgn}\left(-|\pi^{B1, B}_i - \pi^{B1, A}_i| + |\pi^{B2, B}_i - \pi^{B2, A}_i|\right)$ captures the preference for more equal outcomes between the players.

\item $\text{social\_welfare}_i = \text{sgn}\left((\pi^{B1, B}_i + \pi^{B1, A}_i) - (\pi^{B2, B}_i + \pi^{B2, A}_i)\right)$ represents the motivation to maximize the total payoff for both players.

\end{itemize}

Based on this setup, we report the regression coefficients $\beta_{I}$, $\beta_{C}$, $\beta_{D}$, and $\beta_{W}$ to quantify the extent to which each of these factors influences the choices reflected in the LLM-generated responses.

\paragraph{Group Identity on Distributional Preferences.} To investigate the influence of group identity on distributional preferences, we extend the previous regression model by including interaction terms:
\begin{equation}
\begin{split}
a_i & = \beta_0 + \beta_{I} \text{self\_interest}_i + \beta_C \text{competition}_i + \beta_D \text{difference\_aversion}_i + \beta_4 \text{social\_welfare}_W + \\
& \gamma_{I} \text{ingroup} \times \text{self\_interest}_i + \gamma_C \text{ingroup} \times \text{competition}_i + \\ 
& \gamma_D \text{ingroup} \times\text{difference\_aversion}_i + \gamma_W \text{ingroup} \times\text{social\_welfare}_i + 
\epsilon_i
\label{eq:group}
\end{split}
\end{equation}
Here, $\text{ingroup}_i$ is an indicator variable that equals 1 if Player A and Player B share the same group identity in the $i$-th game, and 0 otherwise.
To assess the impact of group identity on distributional preferences, we examine the coefficients for the interaction terms, $\gamma_{I}$,  $\gamma_{C}$,  $\gamma_{D}$, and  $\gamma_{W}$. These coefficients capture how the influence of each factor changes when the players share a group identity.

\paragraph{Direct and Indirect Reciprocity.}
To analyze both types of reciprocity, we calculate the probability of the LLM selecting the option that benefits the other agent more after observing good or bad intentions from others. Reciprocity is then quantified as:
\setlength{\abovedisplayskip}{1pt}
\setlength{\belowdisplayskip}{1pt}
\begin{equation}
 \hat{\mathbb{E}}_{S \in \mathbb{S}^{\text{pro}}} \left[ \hat{\mathbb{E}}_\theta [\mathbbm{1} (A  = a^{\text{pro}}) | S \in \mathbb{S}^{\text{pro}}] \right] - \hat{\mathbb{E}}_{S \in \mathbb{S}^{\text{non-pro}}} \left[ \hat{\mathbb{E}}_\theta [\mathbbm{1} (A  = a^{\text{pro}}) | S \in \mathbb{S}^{\text{non-pro}}] \right], 
\end{equation}
\noindent where $\mathbbm{1}(\cdot)$ is the indicator function, and $a^{\text{pro}}$ represents the prosocial action that benefits the other agent more. The sets $\mathbb{S}^{\text{pro}}$ and $\mathbb{S}^{\text{non-pro}}$ refer to states in which the matched agent (player A) has engaged in prosocial or non-prosocial behavior, respectively.

This equation captures how the previous actions of the matched agent influence the LLM's likelihood of selecting prosocial actions in subsequent rounds. If the LLM is more likely to generate prosocial actions following prosocial behavior from the match, or less likely following non-prosocial behavior, this indicates reciprocity. The magnitude of the change in this probability—by comparing conditions of the match's previous good versus bad actions—quantifies the strength of reciprocity exhibited in the LLM's responses.

Experiment results for distributional preferences and reciprocity are presented in Sections~\ref{subsec:distributional} and Sections~\ref{subsec:reciprocity}, respectively. Prompt sensitivity analysis is presented in Section~\ref{subsec:heterogeneity}.

\subsection{Probabilistic Modeling of Stated Values ($V$) in Reasoning and Action ($A$)}
\label{subsec:prob_reasoning}
The CoT processes provided by LLM are essentially a form of qualitative data, similar to interviews or free response surveys. 
Analyzing CoT processes enables us to understand how and why LLMs derive their final decisions.

\subsubsection{Stated Value Extraction via Deductive Coding.}
\label{subsebsec_deductive_coding}
We adopt a classical qualitative analysis approach in social science known as \textit{deductive coding} to extract stated values~\citep{linneberg2019coding}. In this approach, researchers develop a codebook that contains a list of predetermined definitions applied to their data. This method is particularly useful for analyzing the complex behaviors of LLMs, as it allows us to systematically categorize and interpret their responses. 
There is growing literature demonstrating the effectiveness of LLMs in performing deductive coding~\citep{chew2023llm, tai2024examination}.

Our codebook, denoted by $\mathcal{C}$, consists of defined mechanisms and illustrative examples that highlight the reasoning behind these mechanisms.  To construct the codebook, we first survey the literature to identify the underlying mechanisms governing specific forms of social preference (e.g.,~\cite{andreoni1990impure,fehr1999theory,bolton2000erc,fehr2000fairness}). We then include definitions of these concepts and provide one-shot examples explaining how each concept can be applied in our experimental setup.\footnote{This is consistent with few-shot prompting techniques~\citep{brown2020language}.}

LLMs excel in identifying patterns in textual data. This pattern recognition capability is crucial for deductive coding, where patterns related to specific theoretical concepts or categories need to be consistently recognized across varied data sets. LLMs can help map patterns to predefined codes (stated values) in our codebook, thereby facilitating an accurate and consistent coding process.
Moreover, LLMs can process a substantial amount of data rapidly, which makes them highly efficient tools for deductive coding.\footnote{{After deductive coding with another LLM that is not examined in our paper, we randomly sample 100 reasonings and ask a research assistant to verify and ensure the deductive coding quality.}}
We present our prompt for deductive coding in Appendix~\ref{appendix_deductive_coding}. 

Formally, for a given prompt or state ($S$) and an LLM $\theta$, the model generates a response $R$, which is a sequence of tokens ($R = x_{T_{\text{EOP}+1}:T}$). As previously mentioned, we partition the entire response text $R$ into two parts: $R = (R^{\text{cot}}, R^{\text{act}})$. The CoT part ($R^{\text{cot}}$) consists of a sequence of sentences $R^{\text{cot}} = (R_1, R_2, \dots, R_{n_i})$, where $n_i$ is the number of sentences in the CoT response. We apply deductive coding on a sentence-by-sentence basis using the codebook $\mathcal{C}$, resulting in a sequence of labels $C_1 = C(R_1), C_2 = C(R_2), \dots, C_{n_i} = C(R_{n_i})$. Each label can be either ``understanding'' ($U$) of the instruction or a type of ``stated value'' ($V$), such as fairness or self-interest. From the final action part of the response ($R^{\text{act}}$), we determine the action as $A = A(R^{\text{act}})$. This process results in a decision path for each response to $S$, represented as a sequence: $(C_{1}, C_{2}, \dots, C_{n_i}, A)$. We present the occurrence probability for each stated value in Section~\ref{subsec:occ}.

\subsubsection{Predictability of Stated Values Towards Decisions.}

A central question when analyzing LLMs' responses revolves around whether CoT   reasoning is analyzable and whether these CoT outputs are valid and consistent with the final actions. If we cannot establish the legitimacy of analyzing CoT, there would be no justification for such analysis~\citep{shiffrin2023probing}.

To address this concern, we frame the problem as a prediction task: Given a prompt or state $S$ and the CoT portion of the response $R^{\text{CoT}}$, can we accurately predict the final action $A$? Specifically, we model the following conditional probability for an LLM parameterized by $\theta$: 
\begin{equation} 
\mathbb{P}_\theta(A \mid R^{\text{cot}}, S).\label{eq:pred_cot} 
\end{equation}

If this conditional probability can be effectively modeled using a standard, 
machine learning prediction model (such as XGBoost by \cite{chen2016xgboost}), we contend that the CoT processes are both meaningful and relevant to the final actions. In such a case, the CoT can be interpreted as representing reasoning patterns similar to human responses in surveys or interviews.

The deductive coding in Section~\ref{subsebsec_deductive_coding} provides one way to engineer features from the CoT text $R^{\text{cot}}$. Thus, we use a machine learning prediction algorithm to model the conditional probability distribution in Eq.~\eqref{eq:pred_cot} by utilizing the following features:
\begin{itemize}
    \item \textit{Pre-defined values in the codebook}: For each value in the codebook $\mathcal{C}$, we create a feature that examines whether it is mentioned in the labels of the response $C_1, C_2, \dots$. If so, the feature takes the value 1; otherwise, it takes 0.
    \item \textit{Parameters of prompt instruction}: This includes parameters ($\pi^{(\cdot)}$)  mentioned in Section~\ref{subsec:prob_action}.
\end{itemize}

For each prompt $S_i$ and its corresponding CoT component of the response $R^{\text{cot}}_i$, we predict an action $\widehat{A_i}$. To assess how accurately these predicted actions match the actual actions $A_i$, we examine the frequency of cases where $A_i = \widehat{A_i}$ across all prompt-response pairs $i$. This frequency forms the \textit{prediction accuracy}, which quantifies how well the stated values revealed through the CoT processes can predict the final actions.
Since classification methods can be sensitive to the choice of classification thresholds (e.g., in binary classification, varying the threshold for ``positive" predictions can shift the balance between true positives and false positives), we also evaluate the model's performance using the \textit{Area Under the Receiver Operating Characteristic Curve (AUC)}, which quantifies the model's ability to discriminate between classes across all possible thresholds.  

This analysis reveals the extent to which the patterns in LLM-generated CoT sequences are consistent with the final actions, indicating whether analyzing these CoT outputs is meaningful for interpreting the models’ response patterns. 
A high level of predictability suggests that the LLMs' final outputs are probabilistically dependent on the stated values expressed in the CoT, rather than generating reasoning superficially like ``stochastic parrots." High predictability enables us to evaluate the reliability of LLM outputs and determine whether the stated values can be used to accurately understand and interpret the final actions $A$. We empirically demonstrate this predictability in Section~\ref{subsec:prediction}.

\subsubsection{Interpreting the Reasoning Processes via Tree-Based Visualization.}
\label{subsubsec:tree_based}

Understanding the sequential patterns in the LLMs' CoT outputs is crucial for interpreting how the models generate their final responses. We propose a probabilistic tree-based approach to visualize and analyze the decision paths reflected in the LLM-generated outputs. For each response to a state $S$, we consider the sequence of labels plus the final action $(C_{1}, C_{2}, \dots, A)$.

To capture the stochastic nature of the LLM's responses and maintain the model's true latent probability distribution, we set the sampling temperature to be 1.0 and generate a large number of responses under the same prompt/state $S$. This ensures that the variability in the model's outputs is preserved, reflecting the inherent probabilistic behavior of the LLM. By aggregating and merging overlapping decision paths across multiple responses, we construct a probabilistic decision tree $\mathcal{T}(S)$, given each state (prompt, $S$) which has: 

\begin{itemize}
    \item Nodes: representing either ``understanding'', a stated ``value'', both represented by $C$ or an action. Leaves are actions ($A$). 
    \item Edges: representing the transitions (conditional probabilities) among different $C$ (and also $A$). 
\end{itemize}

The overall probability of a particular decision path $(C_{1}, C_{2}, \dots, A)$ is given by the product of the conditional probabilities along the path:
\[
\mathbb{P}_\theta(A, C_{1}, C_{2}, \dots | S) =\mathbb{P}_\theta(C_1|S) \left( \prod_{j=2}^{n_i} \mathbb{P}_\theta(C_{j} \mid C_{1:j-1}, S)  \right) \mathbb{P}_\theta(A \mid C_{1:n_i}, S).
\]
Here, each conditional probability represents a node in the decision process, where preceding nodes are treated as conditions. Empirically, this is computed as the fraction of a specific stated value (or understanding), $C_j$, that appears given a previous LLM-generated sequence of stated values (or understanding), $C_{1:j-1}$, and the given prompt $S$.
The last term, $\mathbb{P}_\theta(A \mid C_{1:n_i}, S)$, corresponds to a leaf on the tree, showing how different CoT patterns change the probability of choosing one final action. 
Essentially, this tree groups CoT processes that share similar decision paths.

This probabilistic decision tree provides a visual and mathematical representation of the LLM's CoT processes, allowing us to qualitatively understand the likelihood of certain CoT paths leading to specific actions.
Algorithm~\ref{alg:build_tree} in \textit{Appendix~\ref{appendix:tree_based}} outlines the  procedure for constructing the probabilistic decision tree from the coded responses. See Section~\ref{sub:tree} for examples of tree visualization.

\subsubsection{Probabilistic Dependency between Stated Values and Decisions.} 
The tree-based approach effectively visualizes and qualitatively analyzes how the CoT sequences are associated with final actions. Here, we aim to identify which stated values matter statistically significantly—meaning that mentioning them in CoT processes will change the probability of choosing a final decision. We refer to this method as ``probabilistic dependency analysis.'' 

Formally, for each response indexed by \( i \) with its corresponding final action \( a_i \), and for each possible stated value \( v^{(k)} \), such as fairness, in the codebook \( \mathcal{C} = \{v^{(1)}, v^{(2)}, \dots\} \), indexed by \( k \), we check if \( v^{(k)} \) exists in its sequence of labels \( \{C_1, C_2, \dots \}\). If so, we define a variable \( w^{(k)}_i = 1 \); otherwise, \( w^{(k)}_i = 0 \). In this way, for each \( i \), we construct a value feature vector \( \mathbf{W}_i = [w^{(1)}_i, w^{(2)}_i, \dots] \), where the dimensionality is the number of possible stated values and each element indicates whether a value is mentioned in the current response. 

We then perform a fixed-effects regression model to identify the effect of mentioning a stated value on the final action. For given prompt-response pair (indexed by $i$), this is equivalent to estimating the conditional probability via the following linear probability model, 
\[
\mathbb{P}_\theta(A = a_i \mid C_{1:n_i}, S_i) 
\approx 
\mathbb{P}_\theta(A = a_i \mid \mathbf{W}_i, S_i)  \approx  \boldsymbol{\phi}^\top \mathbf{W}_i + \eta_{s_i},
\]
where \( S_i \) and \( a_i \) represent the prompt and the final action in a given prompt-response pair, and \( \boldsymbol{\phi} \) denotes the vector of regression coefficients. \( \eta_S \) refers to the fixed effect for a specific prompt (holding game contexts and parameters constant). 

Although a specific prompt setup  $S$  can influence both the values expressed in the response $W_i$  and the final actions $a_i$, the fixed-effects model accounts for these exogenous factors. Consequently, it allows us to identify the extent to which mentioning a value in the response affects the probability of generating a particular final action. This relationship can be considered \textit{causal} because, after controlling for exogenous factors, the variation arises from the stochastic CoT process driven by non-zero temperature.

By using this model, we quantify the influence of each stated value on the LLM's generation of final actions, providing a mathematical framework for interpreting the internal CoT and utterance generation process in social interaction settings. By identifying values with statistically significant dependencies on the final actions, we enhance the interpretability and transparency of the reasoning process. See Section~\ref{sub:dependency} for our results for such dependency.

\section{Results: Evaluating LLM Final Actions}
\label{sec:main}

In this section, we analyze LLMs' final actions to evaluate their distributional and reciprocal preferences using the method introduced in Section~\ref{subsec:prob_action}. 
For each model and parameter configuration, we perform five replications with a temperature of 0.2.
We evaluate multiple proprietary and open-source LLMs developed by various organizations, or across different versions from the same firm. 
This variety enables us to examine how model family and capacity might influence their responses. In essence, LLMs generate text by generating the next word or ``token'' in a sequence, based on the preceding context and pre-trained text probability distributions. The model processes prior interactions with a user and generates responses to our experimental ``prompts.'' By comparing across these models, we aim to identify potential factors that could affect LLM preferences.
For proprietary models, we include GPT-3.5 and GPT-4 developed by OpenAI, updated in January and April 2024, respectively. We also incorporate four versions of the LLaMA model by Meta, including LLaMA 2 (13 billion and 70 billion parameters) launched in February 2023, and LLaMA 3 (8b and 70b parameters) launched in April 2024. Additionally, we use Mistral AI models: Mistral 7B (7.3 billion parameter model that is claimed to outperform LLaMA (2-13B)) and Mixtral 8x7 (46.7 billion parameters; claimed to match or outperform GPT 3.5).

To assess distributional preferences, we run each model 1,200 times. For group identity effects in the main results, we replicate each model and parameter configuration five times with a temperature of 0.2, applying CoT for both ingroup and outgroup conditions.\footnote{Robustness checks on temperature and existence of CoT are presented in Section~\ref{sub:robust:cot}.} Following the four methods of inducing group identities—we execute each model 9,600 times.
For both direct and indirect reciprocity, each model is run 1,200 times, covering instances where LLMs are informed that their matches displayed both good intentions and misbehavior.

\subsection{Distributional Preferences}
\label{subsec:distributional}

\begin{figure}
    \centering
    \includegraphics[width=.8\linewidth]{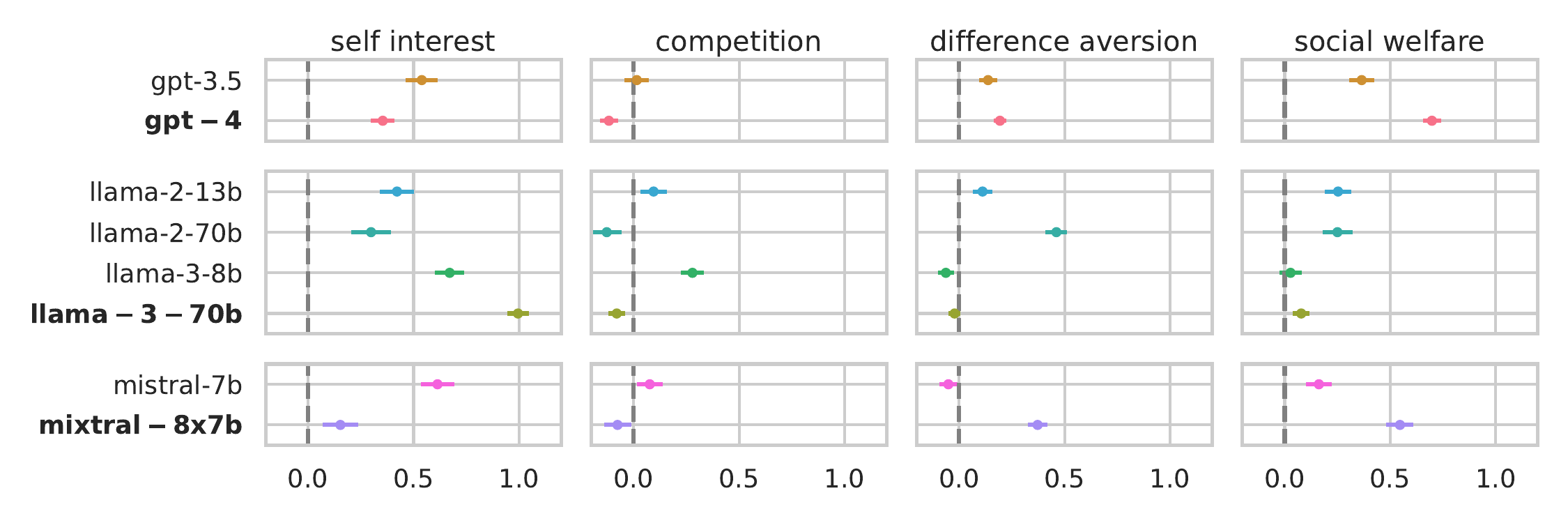}
    \caption{[Main Result 1] Distributional preferences indicated by self interest, competition, difference aversion, and social welfare, reflected by regression coefficients. Error bars are 95\% CIs. }
    \label{fig:distributional}
\end{figure}

In our examination of distributional preferences inferred from actions revealed by LLMs in our lab settings (using regression specified in Eq~\eqref{eq:distribution}), as depicted in 
Figure~\ref{fig:distributional}, several key observations emerged. 
Notably, most models do not generate decisions aligned solely with self-interest (self-interest $<1.0$);
Instead, many model responses state a moderate inclination toward social welfare,  with the exception of LLaMA (2-13B). 
Additionally, the inclination stated in LLM responses toward competition is generally low (close to 0) across most models. While some models, such as LLaMA (2-70B) and Mistral 8x7B, state strong (about 0.4-0.5) tendency for difference aversions, this tendency is overall less pronounced compared to their tendencies toward social welfare or self-interest.

For GPT and Mistral models, there is a trend where increased model capacity (e.g., a larger parameter size or a newer version) correlates with a decline in self-interested behavior and a rise in social welfare-oriented behavior. We hypothesize that this shift may be due to the models' improved ability to align with human preferences from training and alignment processes. As these models grow in complexity, their decision-making appears to become less associated with self-interest, potentially reflecting a deeper understanding of the underlying sophisticated social behavior, such as cooperation and altruism.

Interestingly, in the case of LLaMA models, increased model capacity is associated with a stronger drive toward self-interest. We suspect this may be because the alignment procedures for LLaMA models make them more goal-oriented and utility-maximizing in their decision-making processes~\citep{wang2024comprehensive}. This might lead to decisions that prioritize self-interest, especially in cases where such behavior is more efficient or optimal according to the models' internal logic.

\paragraph{Group identity effect on distributional preferences.}
Following the literature on distributional preferences, we investigate an important moderator for these preferences—group identity. Specifically, we explore whether shared attributes between the match and the player influence  distributional preferences reflected in LLM responses. To analyze this, we conduct the regression specified in Eq.~\eqref{eq:group} and report the interaction effects between group identity and various distributional preferences.

\begin{figure}
    \centering
    \includegraphics[width=.8\linewidth]{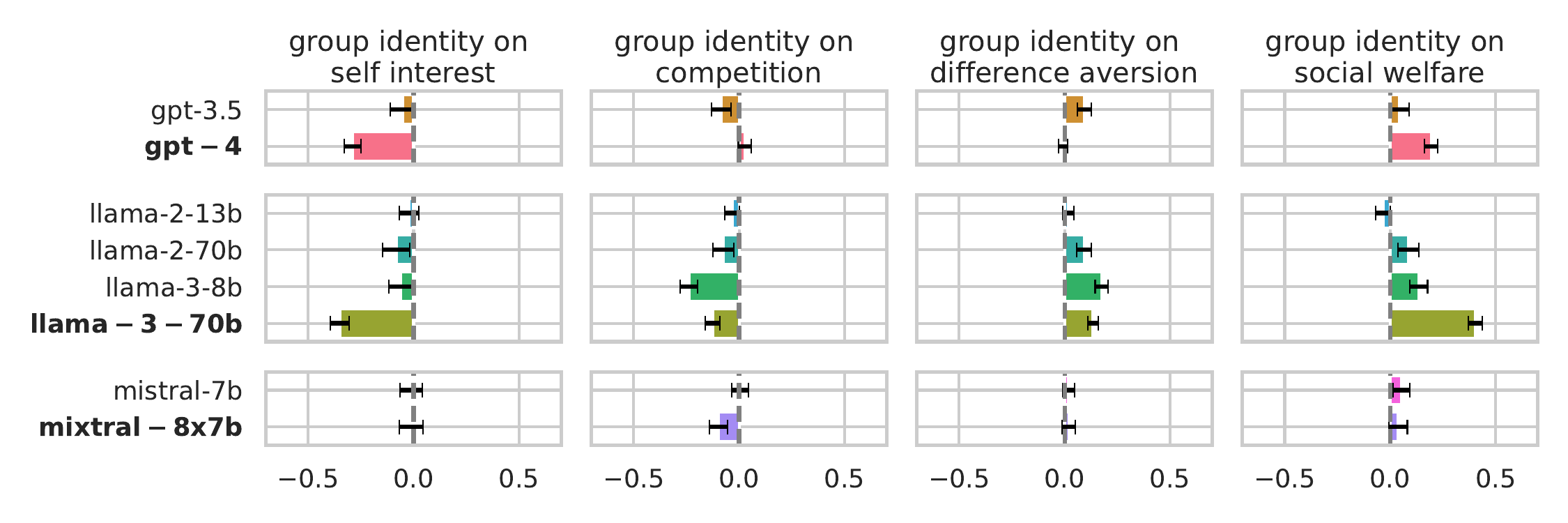}
    \caption{[Main Result 2] Interaction effects of shared group identity and  distributional preferences. Error bars are 95\% CIs. }
    \label{fig:group}
\end{figure}

Figure~\ref{fig:group} illustrates the interaction effects of shared group identity and distributional preferences. First, not all models do exhibit substantial group identity effects, with the exceptions of the two strongest models--GPT-4 and LLaMA (3-70B). These two models generate responses showing a substantial increase in social welfare tendency ($\gamma_W=0.19$ and $0.40$, $p<0.01$) when informed of interacting with ingroup matches, accompanied by a large decrease in self-interest. Second, for some other models, we observe smaller effects; for instance, responses in LLaMA (2-70B) and LLaMA (3-8B) indicate slightly more social welfare and difference aversion ($p<0.01$), and less self-interest and competition when under ingroup conditions ($p<0.05$).

\subsection{Reciprocity}
\label{subsec:reciprocity}
\begin{figure}
    \centering
    \includegraphics[width=0.4\linewidth]{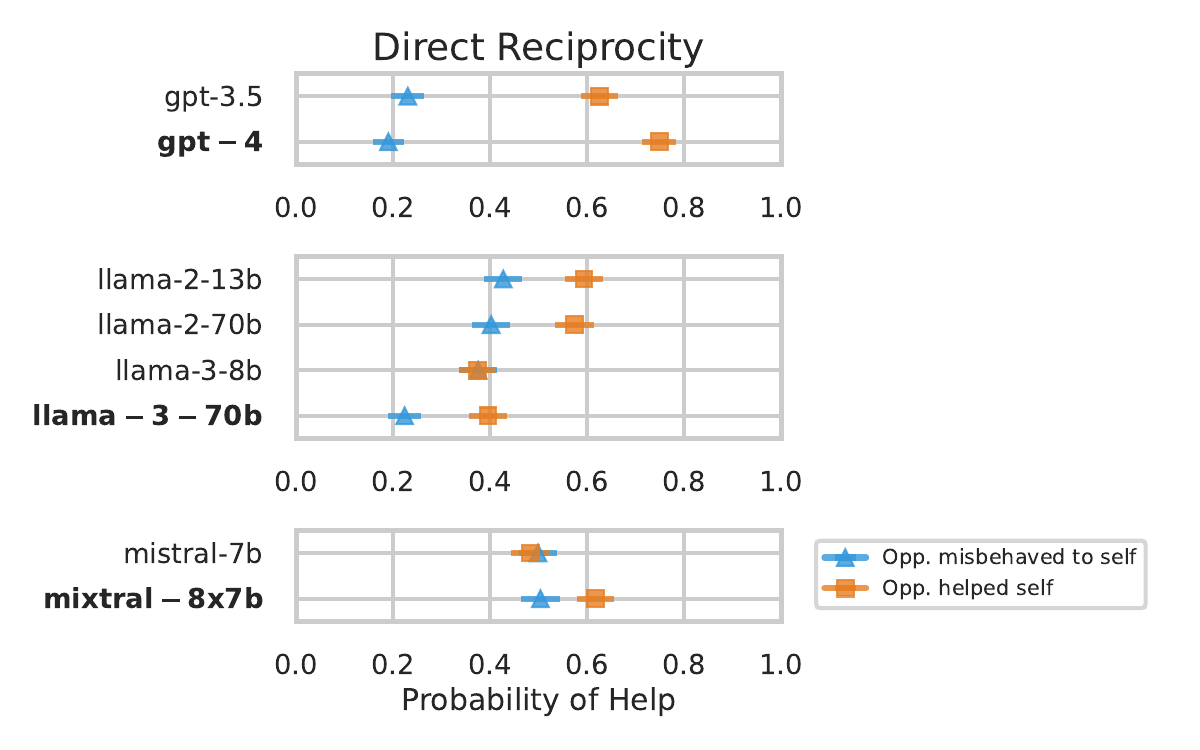}~
    \includegraphics[width=0.4\linewidth]{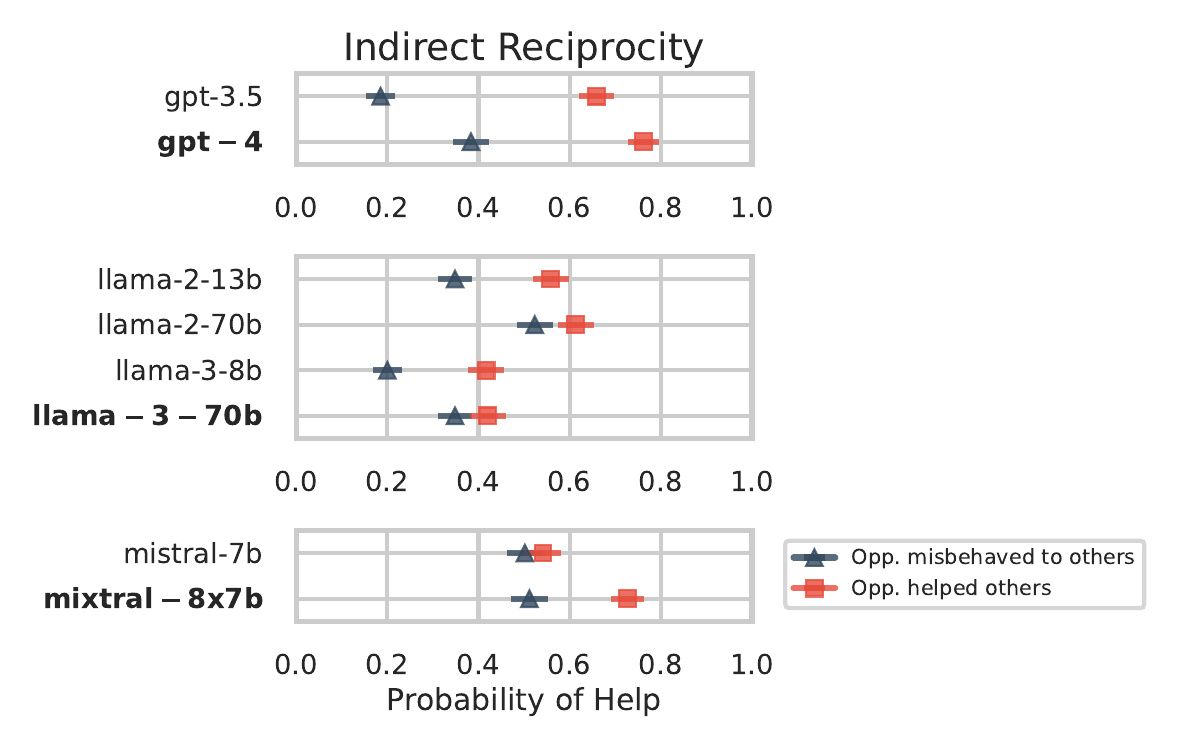}
    \caption{[Main Result 3] Reciprocity preferences for LLMs. The left and right panels represent direct and indirect reciprocity preferences, respectively. The error bars indicate the probability of the LLM behaving prosocially when informed that the match (opponent) has previously helped or misbehaved towards them (direct) or others (indirect). Error bars are 95\% CIs. }
    \label{fig:reciprocity}
\end{figure}

Next, we examine direct and indirect reciprocity in Figure~\ref{fig:reciprocity}.  
From the figure, we observe several findings. First, all models, except for Mistral 7B, exhibit both direct and indirect reciprocity, as evidenced by the differences in their responses under the two conditions related to the match's historical actions. Mistral 7B, however, does not appear to display statistically significant reciprocal behavior ($p>0.1$), which might be due to its limited reasoning ability and its lack of theory of mind~\citep{strachan2024testing} to recognize intentions associated with the match's prior actions.

In addition, responses from the two GPT models indicate the highest levels of reciprocity. They significantly adjust responses based on the historical behavior of their matches, indicating a strong sensitivity to their matches' intentions and thus exhibiting robust, human-like reciprocal behavior.

Responses generated by LLaMA models and Mixtral (8x7B) appear to display numerically smaller reciprocal preferences compared to the GPT models. However, when a match displayed good intentions rather than misbehavior, the likelihood of these LLMs exhibiting positive intentions increases by up to 20 percentage points—a still significant effect.

Interestingly, unlike humans—who typically show a stronger preference for direct reciprocity~\citep{nowak2005evolution} (valuing actions done directly to themselves more than actions done to others)—multiple LLMs, including GPT-3.5, LLaMA (3-8B), and Mixtral (8x7B), display similar levels of both indirect and direct reciprocity (some models have stronger direct reciprocity than indirect whereas the others display the opposite).  This suggests that, unlike humans, LLMs are trained or aligned in a way that does not prioritize actions involving them directly over those involving others. In other words, LLMs may be trained or designed to avoid the human tendency to prioritize personal interactions, processing reciprocal actions in a more neutral and consistent way, regardless of whether the actions are directed toward them or others.

\subsection{Additional Analyses}
\label{subsec:heterogeneity}
A major challenge in contemporary research of understanding LLMs' behavior is the proper design of prompts. Within our \textit{SUVA} framework, this involves defining a set of environmental states $\mathbb{S}$ that capture general LLM responses in simulated social interactions.
In this section, we perform sensitivity analyses to evaluate how variations in prompt design $S$ as well as temperature, affect our findings. Due to space constraints, we focus on the most capable models from each family: GPT-4, LLaMA (3-70B), and Mixtral (8x7B). The full analysis are provided in Appendix~\ref{Appendix:Heterogeneity_robustness}. 

\subsubsection{Sensitivity to Incentive Structure.}
To examine the effect of incentive structure, we vary two factors in the system prompt: income levels and incentive conversion rates. The income levels correspond to the 10th, 50th, and 90th percentiles in the 2023 U.S. labor market, with annual incomes of \$17,000, \$74,202, and \$216,056, respectively. The incentive conversion rate determines the monetary value per 100 points earned, with conversion values of \$0.01, \$1, \$100, and \$10,000.

\begin{figure}
    \centering
    \includegraphics[width=.9\linewidth]{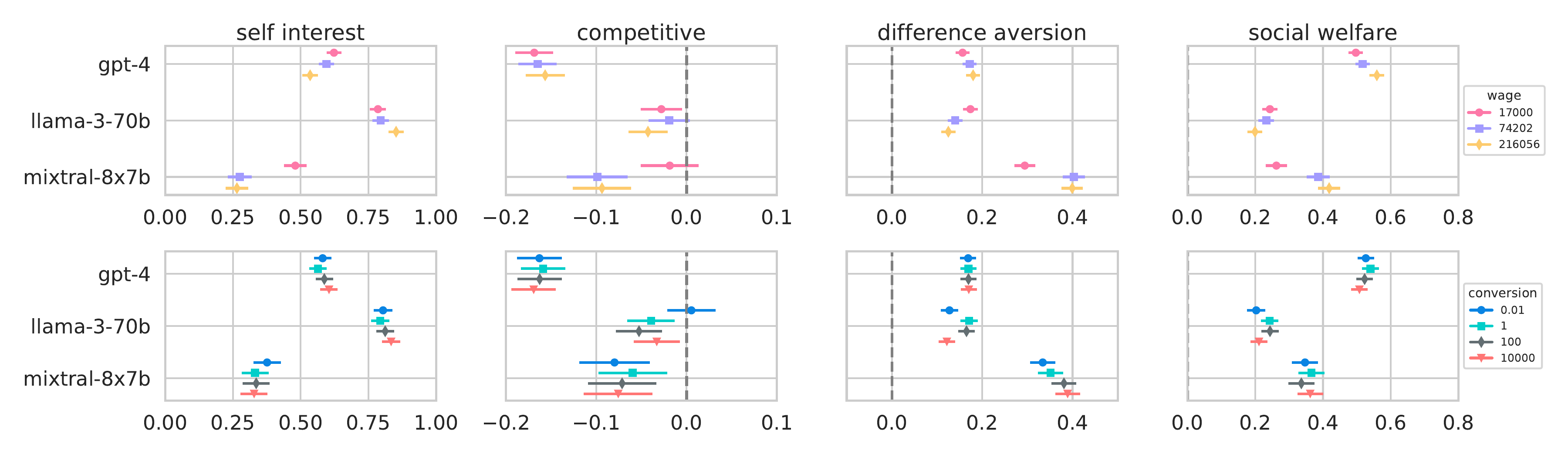}
    \caption{Prompt sensitivity analysis for incentive structures. This analysis explores the sensitivity of distributional preferences to different incentive structures, i.e., changes in wages and conversion rates in the system prompts. Error bars represent 95\% CIs.
    }
    \label{fig:incentive}
\end{figure}

As shown in Figure~\ref{fig:incentive}, social preferences reflected in different models are generally not very sensitive to changes in incentive structures, including variations in conversion rates and income levels (which alter the relative value of additional gains). Across all experimental conditions, self-interest, competition, difference aversion, and social welfare remain largely consistent, with no substantial changes observed as the incentive structure shifts (see Appendix~\ref{appendix:incentive} for detailed discussion).\footnote{{The only noticeable difference occurs when income is at a low level (\$17,000); at this point, GPT-4 and Mixtral (8x7B) appear slightly more self-interested compared to higher income levels.}}

\subsubsection{Persona Effects.}

Next, we investigate the impact of persona variables—an effective approach to tailoring LLMs to specific contexts—by examining whether including demographic and behavioral factors impact LLM responses. The personas are designed based on the dataset from \cite{chen2009group}, incorporating demographic information and behavioral data such as donation or volunteering experience, as well as strategy preferences such as ``\textit{try to maximize own payoff}''. Specifically, for each replicate in our experiment, we randomly select a participant from \cite{chen2009group} and apply their demographic and strategic personas accordingly. This approach ensures that the participant pool's distribution from \cite{chen2009group} is accurately represented, enabling us to examine both heterogeneity and aggregate outcomes.

\begin{figure}
    \centering
    \includegraphics[width=.87\linewidth]{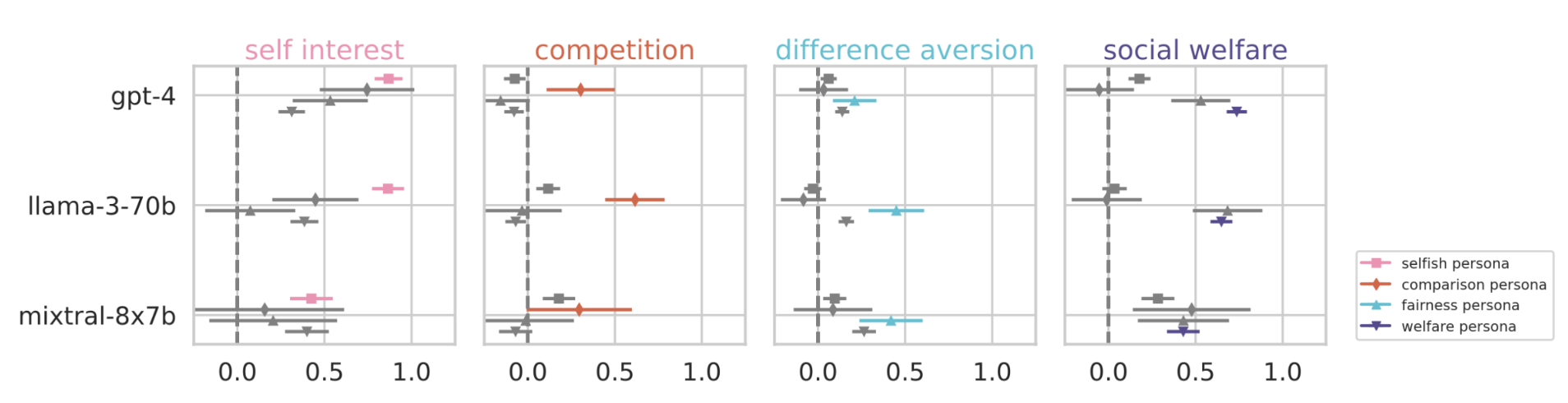}
    \caption{Prompt sensitivity analysis for personas. This analysis explores the sensitivity of distributional preferences to different strategic personas. Error bars represent 95\% CIs.}
    \label{fig:persona}
\end{figure}

We primarily focus on strategic personas where heterogeneity is evident.\footnote{We also examine demographic factors like age, gender, and donation history, which appear in the human participants dataset in \cite{chen2009group}, but they do not produce significant variability in outcomes. } As shown in Figure~\ref{fig:persona}, our analysis reveals that strategic personas, which are directly related to decision-making strategies, significantly influence LLM responses in distributional preference games.\footnote{Note that a very small portion of participants mentioned competition or fairness thus their confidence intervals are wider than selfish or welfare personas.} LLMs assigned with a specific strategy persona show a stronger tendency towards the corresponding dimension in distributional preferences. For instance, when fairness is mentioned in the system prompt, all three models show the highest tendency toward difference aversion. 

Moreover, despite the diverse personas assigned—each corresponding to a unique participant from \cite{chen2009group}—the aggregated results shown in the bottom row in Figure~\ref{fig:robustness} indicate that aggregate results are qualitatively consistent with the experimental results without personas (our main results). For example, when considering aggregate outcomes with every LLM being assigned with a persona, all three models still exhibit substantial levels of distributional and reciprocal preferences. 
However, the magnitudes change slightly. For instance, the aggregate results indicate higher self-interest for GPT-4 but lower self-interest for LLaMA (3-70B). We hypothesize that this variation is influenced by the proportion of personas assigned to each model. Specifically, selfish and welfare-oriented personas are the most common in the dataset from \cite{chen2009group}, leading to a majority of LLMs being assigned these personas. Sensitivity to persona types may vary across different model families, leading to variations in how the results deviate from the main findings.  See Appendix~\ref{appendix:persona} for detailed analysis.

Overall, these findings suggest that LLM social preferences are not sensitive to generic demographic information but can be guided by prompts that specify certain strategies. These results indicate both the adaptability of our evaluation method to contexts with specific personas and the potential comparability between contexts with unspecified personas and the ``vanilla'' version where no persona is specified. 

\subsubsection{Temperature and CoT.}
\label{sub:robust:cot}
We conduct further analyses to assess the impact of temperature settings and the inclusion of CoT prompting on the distributional preferences shown in model responses.

\paragraph{Temperature Settings.}

We explore the effects of increasing the temperature from 0.2 (used in the main analysis) to 0.8 (which is close to the default setup of most commercial chatbots). A higher temperature allows for greater variability in the generated text but can also result in less coherent or more random outputs. 

\begin{figure}
    \centering
    \includegraphics[width=0.95\linewidth]{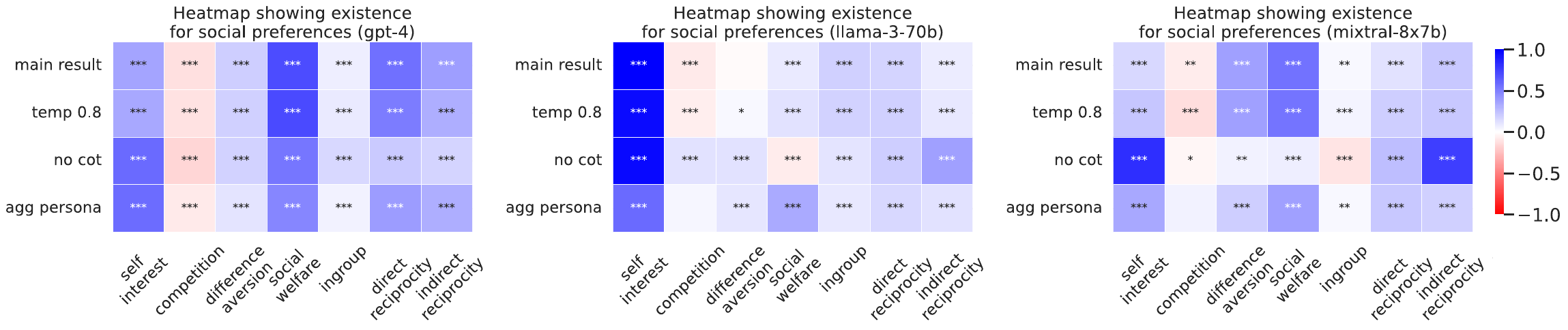}
        \caption{Sensitivity analysis for temperature, the presence of chain-of-thought (CoT), and aggregate results when strategic and demographic personas are assigned. ``Ingroup," ``direct reciprocity," and ``indirect reciprocity" refer to the change in the probability of acting prosocially when the match shifts from ``outgroup" to ``ingroup" or when the counterpart has misbehaved versus shown good intentions. ``Main result" refers to the results presented in Sections~\ref{subsec:distributional}~and~\ref{subsec:reciprocity}. ``Temp 0.8" indicates that the temperature is adjusted from 0.2 to 0.8 relative to the main result. ``No CoT" means the model generates output directly, without intermediate reasoning. ``Agg persona" refers to the aggregate results when strategic and demographic personas are assigned to each replicate. *, **, *** indicate significant levels for $p<0.1$, $p<0.05$, and $p<0.01$ respectively.}
    \label{fig:robustness}
\end{figure}

As shown in the first two rows in Figure~\ref{fig:robustness}, by changing the temperature from 0.2 to 0.8, both evaluation on distributional preferences and reciprocity remains robust. 
This result indicates that temperature adjustments have minimal effects on their decision-making processes (detailed results are provided in Appendix~\ref{appendix:robustness}).

\paragraph{Chain-of-Thought Prompting.}

We examine the impact of removing CoT prompting, specifically the instruction to ``\textit{Please reason step-by-step}." The results by comparing the first and third rows in Figure~\ref{fig:robustness} show that CoT prompting can influence the social preferences reflected in model responses. 
While CoT reasoning generally prompts the model to respond reflecting more prosociality in some models (GPT-4 and Mixtral (8x7B) precisely), the effects are model-specific.\footnote{{For instance, CoT appears to prompt LLaMA (3-70B) to exhibit more deliberative and quantitative thinking, which leads to more self-interested behavior. }} These findings underscore the importance of considering model characteristics and application contexts when deciding whether to include CoT prompting. See Appendix~\ref{appendix:robustness} for detailed analysis.


\section{Results: Tracing Utterance-Based Reasoning to Final Actions}
\label{sec:mechanism}
Adopting the common practice of using CoT not only enhances a model's reasoning ability but also helps to better understand why models arrive at certain decisions. Within our \textit{SUVA} framework, the analysis of CoT is equivalent to discussing the probability distribution $\mathbb{P}_{\theta}(A \mid U, V, S)$. In this section, we first focus on making the value $V$ interpretable by providing deductive coding of the sentences generated by LLMs. Next, we examine how the model’s understanding $U$ and the revealed value $V$ lead to final actions $A$.

\subsection{Occurrences of Stated Values $V$ of LLMs}

We apply deductive coding to determine whether specific values from a predefined codebook are discussed in Section~\ref{subsebsec_deductive_coding}. To begin with, we randomly sample a small subset of LLM responses and summarize the common values present. Next, GPT-4o, the state-of-the-art model not included in the main analysis in Section~\ref{sec:main}, is instead employed to label sentences according to the values discussed. If no explicit values are mentioned and the sentence simply repeats the instruction ($S$), the sentence is labeled as ``understanding" the instruction. This labeling process is iteratively refined until the results align with human-coded labels.

In the left panels of Figure~\ref{fig:combined_heatmaps}, we present the occurrence probabilities of each value for different models, ranging from 0 to 1. In the distributional preference scenarios, we observe that self-interest is consistently high across models, reaching its peak at 1.00 for LLaMA (3-8B). Altruism and fairness show moderate values, though they vary across models, while competition consistently displays very low probabilities.

In scenarios involving group identity, self-interest remains high but slightly lower than in the distributional preference experiment. Ingroup cooperation and fairness show moderate values, while outgroup competition and cooperation display low probabilities. 

For reciprocity, self-interest remains prominent but generally lower than in the distributional preference scenarios. Theory of mind and fairness exhibit higher probabilities only in  certain models, and positive reciprocity demonstrates moderate values in models like GPT-4, LLaMA (3-70B), and Mixtral (8x7B). For indirect reciprocity , self-interest remains high but at lower levels than in other experiments, while altruism and positive reciprocity display notable values in some models (e.g. GPT-4, LLaMA (3-70B), and Mixtral (8x7B) for altruism).

\label{subsec:occ}

\subsection{Predictability of Stated Values $V$ Towards Final Actions $A$}

To assess the extent to which stated values and payoff parameters influence final actions of LLMs, we employ a machine learning approach that quantifies the predictability of prosocial decisions using stated values and payoff parameters as features.\footnote{Note that the features employed in our analysis, specifically the stated values derived from our pre-designed codebook, are not exhaustive. Future work could incorporate additional features, such as sentiment analysis, bag-of-words representations, or word embeddings, to potentially improve prediction performance further.}
Specifically, we utilize the Extreme Gradient Boosting (XGBoost) algorithm, a widely-adopted and robust machine learning model, optimized through grid search for hyperparameter tuning to ensure optimal performance.

To mitigate the impact of interdependencies among observations that share identical game parameter setups—which could lead to similar decision outcomes—we implement a careful data splitting strategy. The dataset is randomly partitioned into training and test sets such that all replicates originating from the same game parameter configuration are entirely contained within either the training or the test set.  This helps mitigate overfitting in the prediction tasks.

We report prediction accuracy for each model within each game and also present the AUC to account for the potential imbalance in the dataset.
As presented in Table~\ref{tab:prediction}, the prediction accuracy for most games and models exceeds 90\%, indicating that stated values are sufficient to achieve reasonable prediction performance for final actions. Furthermore, recognizing that accuracy may not fully capture performance in the presence of class imbalance, the consistently high AUC values further validate the model’s strong predictive capabilities.

Additionally, our analysis reveals that more advanced models—such as GPT-4, LLaMA (3-70B), and Mixtral (8x7B)—consistently outperform their less advanced counterparts within the same families in terms of both accuracy and AUC. This suggests that these advanced models exhibit greater consistency between their CoT reasoning and final decisions.

In summary, the high prediction accuracy and AUC values demonstrate that the CoT reasoning processes of LLMs are both meaningful and relevant to their final actions. This supports the legitimacy of analyzing CoT as a valid method for interpreting and predicting LLM behaviors. Moreover, the superior performance of more advanced models underscores the enhanced coherence and transparency in their reasoning and decision-making processes, reinforcing their suitability for applications requiring reliable and interpretable AI decision-making.
\begin{table}[ht]
\centering
\caption{Prediction Performance with Stated Values as Predictors and Action as Outcome}\label{tab:prediction}
\tiny\begin{tabular}{c|cc|cc|cc|cc}
\toprule
\textbf{Model} & \textbf{Accuracy} & \textbf{AUC} & \textbf{Accuracy} & \textbf{AUC} & \textbf{Accuracy} & \textbf{AUC} & \textbf{Accuracy} & \textbf{AUC} \\
              & \textbf{(Distributional)} & \textbf{(Distributional)} & \textbf{(Group)} & \textbf{(Group)} & \textbf{(Direct)} & \textbf{(Direct)} & \textbf{(Indirect)} & \textbf{(Indirect)} \\
\midrule
GPT-3.5 & 0.919 & 0.954 & 0.849 & 0.918 & 0.786 & 0.845 & 0.739 & 0.795 \\
GPT-4 & 0.978 & 0.996 & 0.967 & 0.996 & 0.778 & 0.858 & 0.872 & 0.949 \\
LLaMA (2-13B) & 0.902 & 0.957 & 0.863 & 0.931 & 0.828 & 0.897 & 0.800 & 0.885 \\
LLaMA (2-70B) & 0.825 & 0.884 & 0.794 & 0.852 & 0.842 & 0.930 & 0.853 & 0.933 \\
LLaMA (3-8B) & 0.915 & 0.967 & 0.870 & 0.935 & 0.834 & 0.881 & 0.797 & 0.867 \\
LLaMA (3-70B) & 0.958 & 0.989 & 0.940 & 0.982 & 0.894 & 0.946 & 0.922 & 0.970 \\
Mistral (7B) & 0.859 & 0.915 & 0.841 & 0.916 & 0.745 & 0.828 & 0.858 & 0.921 \\
Mixtral (8x7B) & 0.924 & 0.963 & 0.884 & 0.939 & 0.825 & 0.900 & 0.852 & 0.912 \\
\bottomrule
\end{tabular} 
\end{table}
\label{subsec:prediction}

\subsection{Visualizing Utterance-Based Reasoning via Decision Trees}
\label{sub:tree}

\begin{figure}
    \centering

\makebox[\linewidth]{
\subfloat[Distributional preference  \label{fig:tree_dictator}]{
    \includegraphics[width=0.55\linewidth]{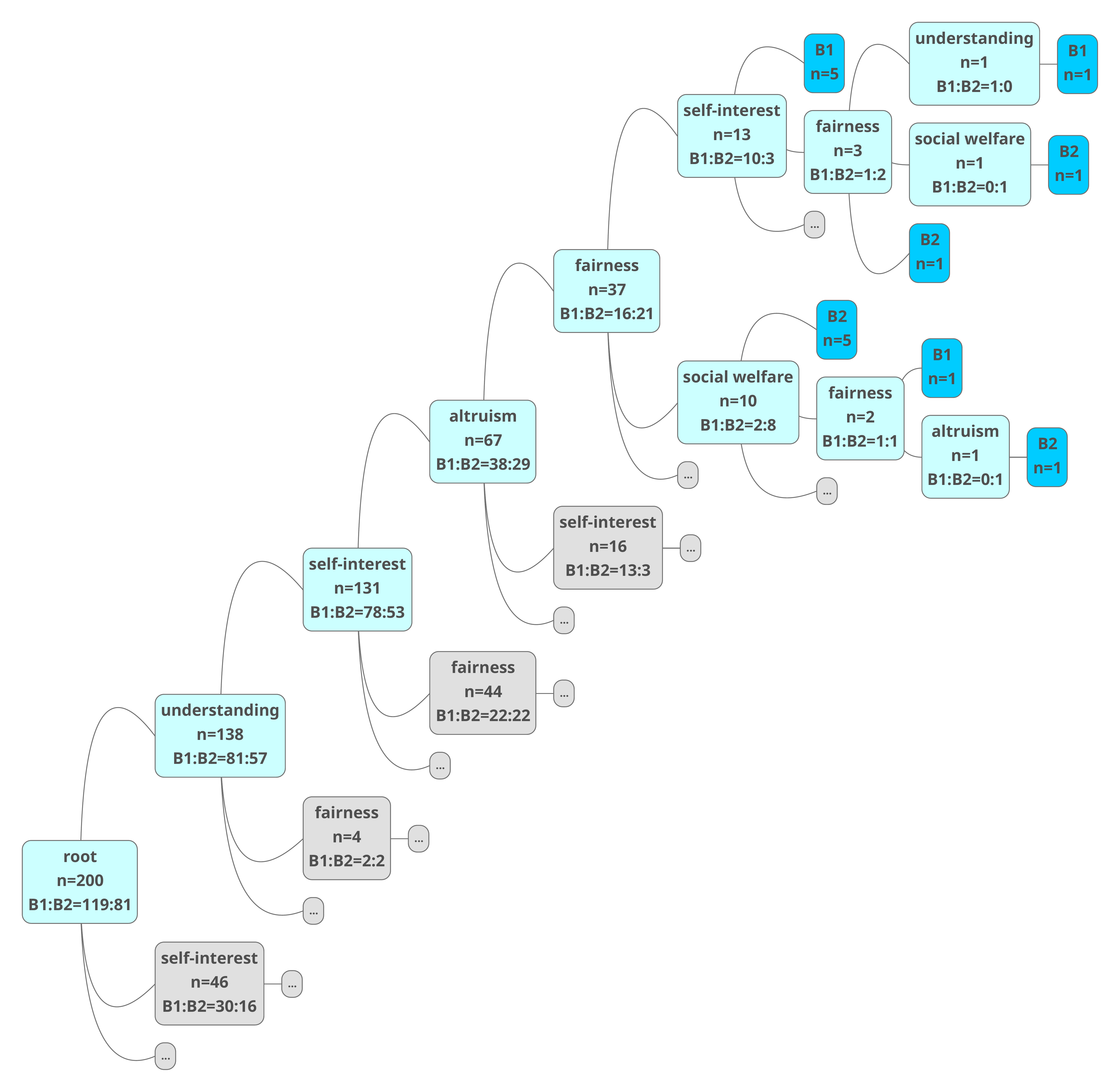}
}
\subfloat[Group identity effect  \label{fig:tree_group}]{
    \includegraphics[width=0.55\linewidth]{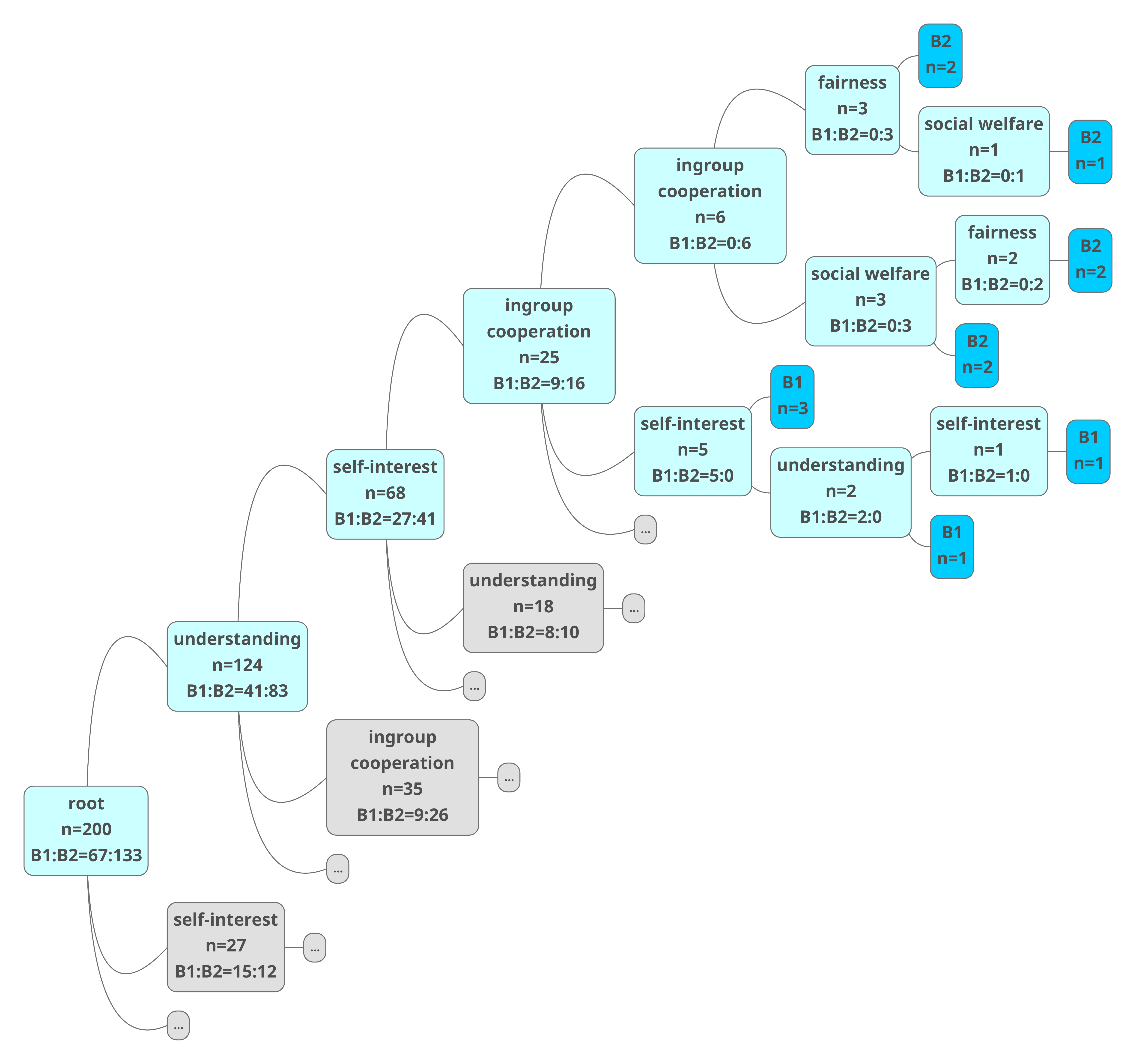}
}}\\
\makebox[\linewidth]{
\subfloat[Direct reciprocity  \label{fig:tree_reciprocity}]{
    \includegraphics[width=0.55\linewidth]{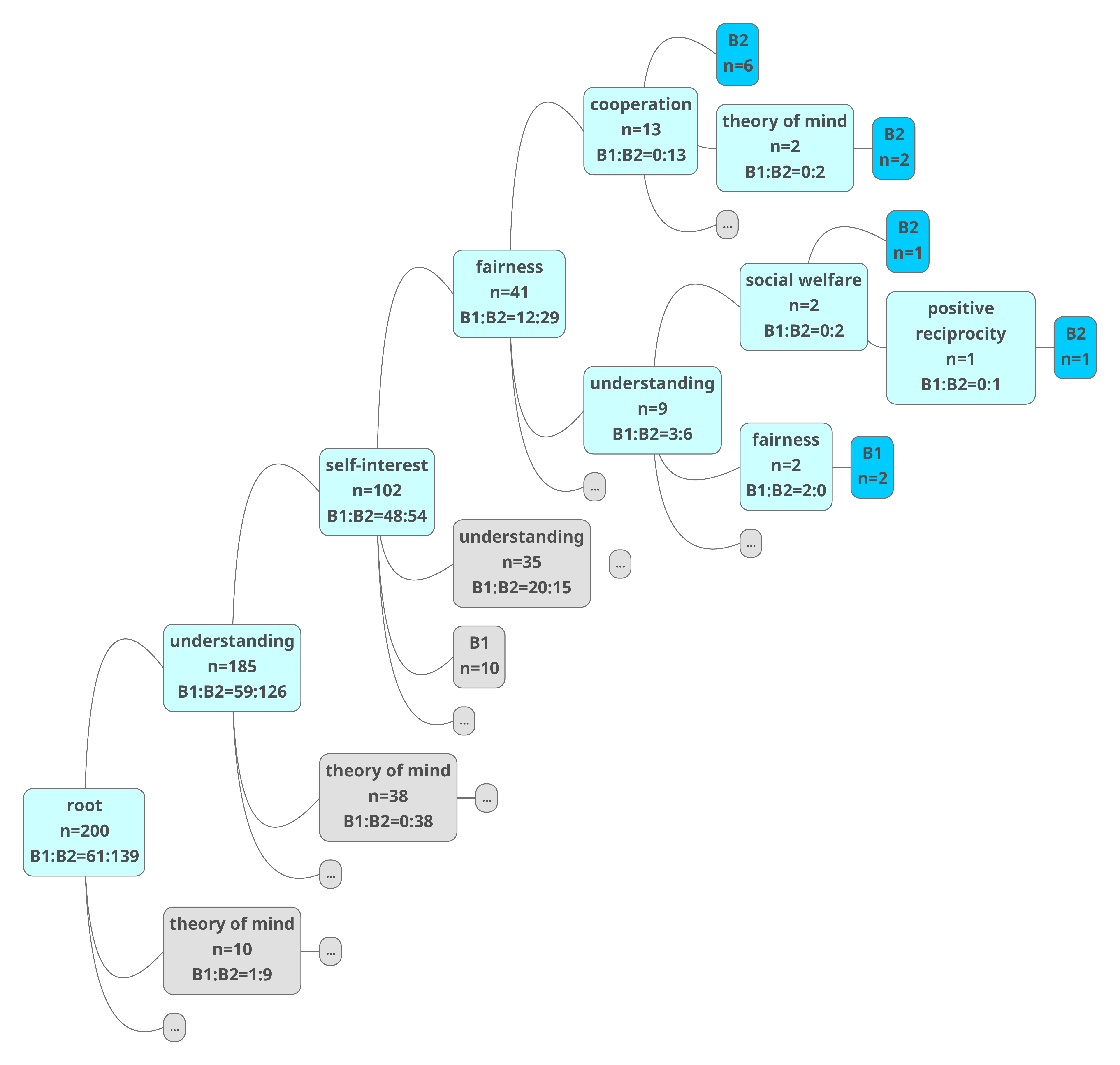}
}
~
\subfloat[Indirect reciprocity  \label{fig:tree_indirect_reciprocity}]{\includegraphics[width=0.55\linewidth]{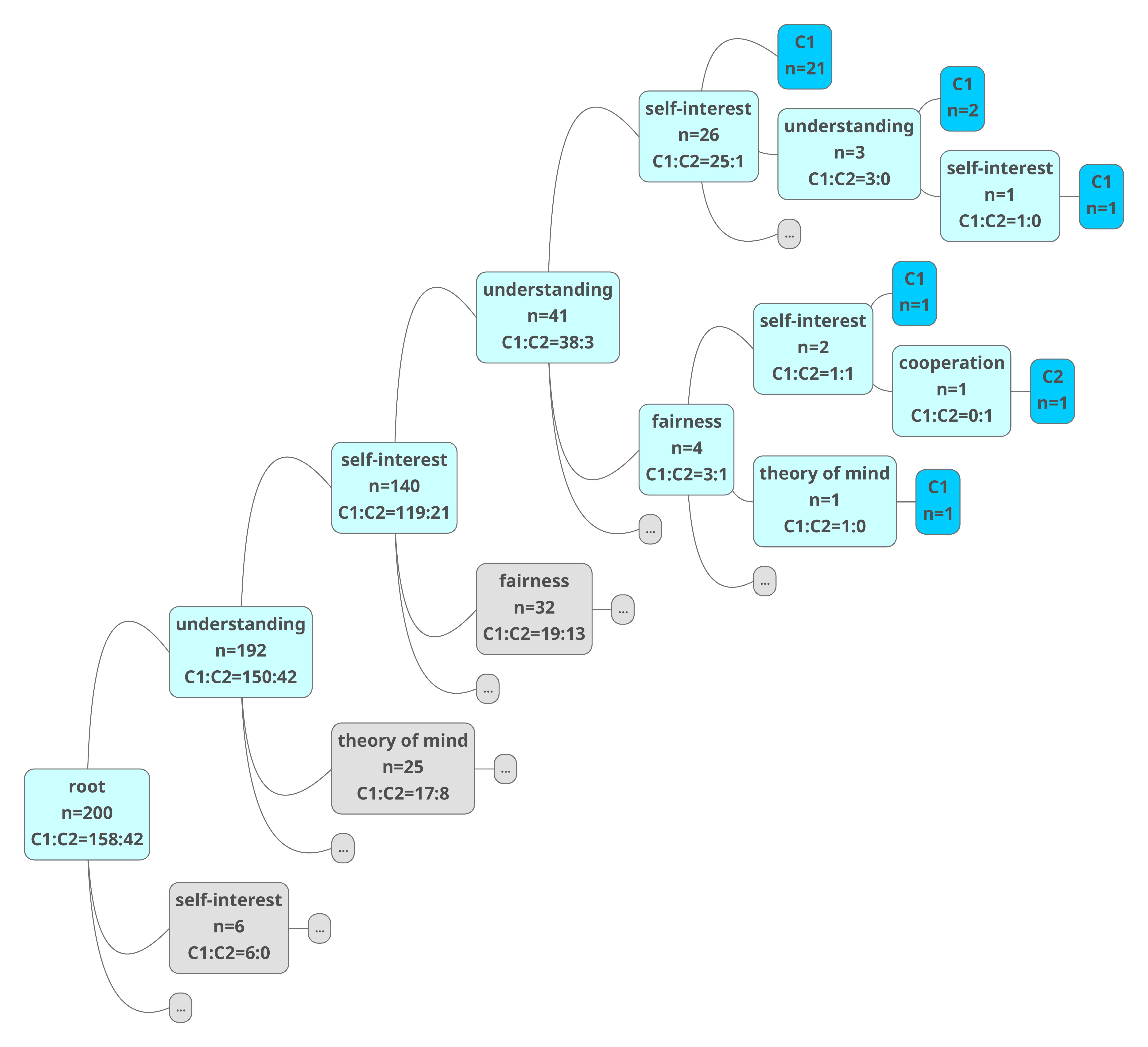}}
}

\caption{Examples of reasoning trees. Payoffs are consistent across all panels, with B1’s (or C1's) payoff = (400, 800) representing a more selfish decision, and B2’s (or C2's) payoff = (800, 600) representing a more altruistic decision. The group identity context uses an example of an ingroup match, while the reciprocity context illustrates a scenario where a match has previously provided help. These results were generated using GPT-4 with 200 replications at a temperature setting of 1, reflecting the true probability distribution of completions given the prompt. The number of replications passing through each decision node is represented by $n$, with the numbers of paths leading to B1/C1 (selfish decision) versus B2/B2 (altruistic decision) also indicated. ``...'' indicates that the branch is folded.  All panels demonstrate that final decisions are influenced by the reasoning paths.
}
    \label{fig:trees}
\end{figure}

Using GPT-4 as an example, we visualize reasoning trees for each experimental context in Section~\ref{sec:main}. Across all settings, we use the payoff structures where B1 = (400, 800), representing a more selfish decision, and B2 = (800, 600), representing a more altruistic decision. The group identity context is illustrated with an ingroup match example.  The direct reciprocity context features a scenario where one party has previously provided help whereas the indirect reciprocity features a scenario where one party has previously misbehaved. The resulting decision trees are presented in Figure~\ref{fig:trees}. Due to space constraints, we show only the top two branches under each decision node and fold the rest of the less likely paths.

As shown in Figure~\ref{fig:tree_dictator}, following the most likely decision path (in light blue), when “self-interest” is revealed, the B1:B2 ratio shifts to favor B1, indicating a more selfish decision. Conversely, when “social welfare” is revealed, the ratio favors B2, reflecting a more altruistic decision. The node preceding the final decision appears to be highly significant—values like social welfare and altruism are followed by B2 in this illustration.

For the group identity effect (Figure~\ref{fig:tree_group}), similar conclusions emerge. When the model mentions ingroup cooperation, the B1:B2
ratio becomes 0:6, whereas “self-interest” leads to a B1:B2
ratio of 5:0. In the reciprocity experiment (Figure~\ref{fig:tree_reciprocity}), we observe that stated values like positive reciprocity and social welfare result in a decision favoring B2, while self-interest as the final node leads to a decision favoring B1.

In conclusion, our tree-based approach provides an effective way to visualize and interpret the decision-making process of LLMs. By mapping decision paths and stated values, we can better understand how models like GPT-4 shift between selfish and altruistic outcomes across different contexts.
This method allows practitioners and researchers to understand patterns in the model's reasoning process, and also emphasizes the probabilistic nature of LLM decision-making.

\subsection{Probabilistic Dependency Analysis Between Stated Values $V$ and Actions $A$}\label{sub:dependency}
\begin{figure}
    \centering
    \makebox[\linewidth]{
    \includegraphics[width=1.23\linewidth]{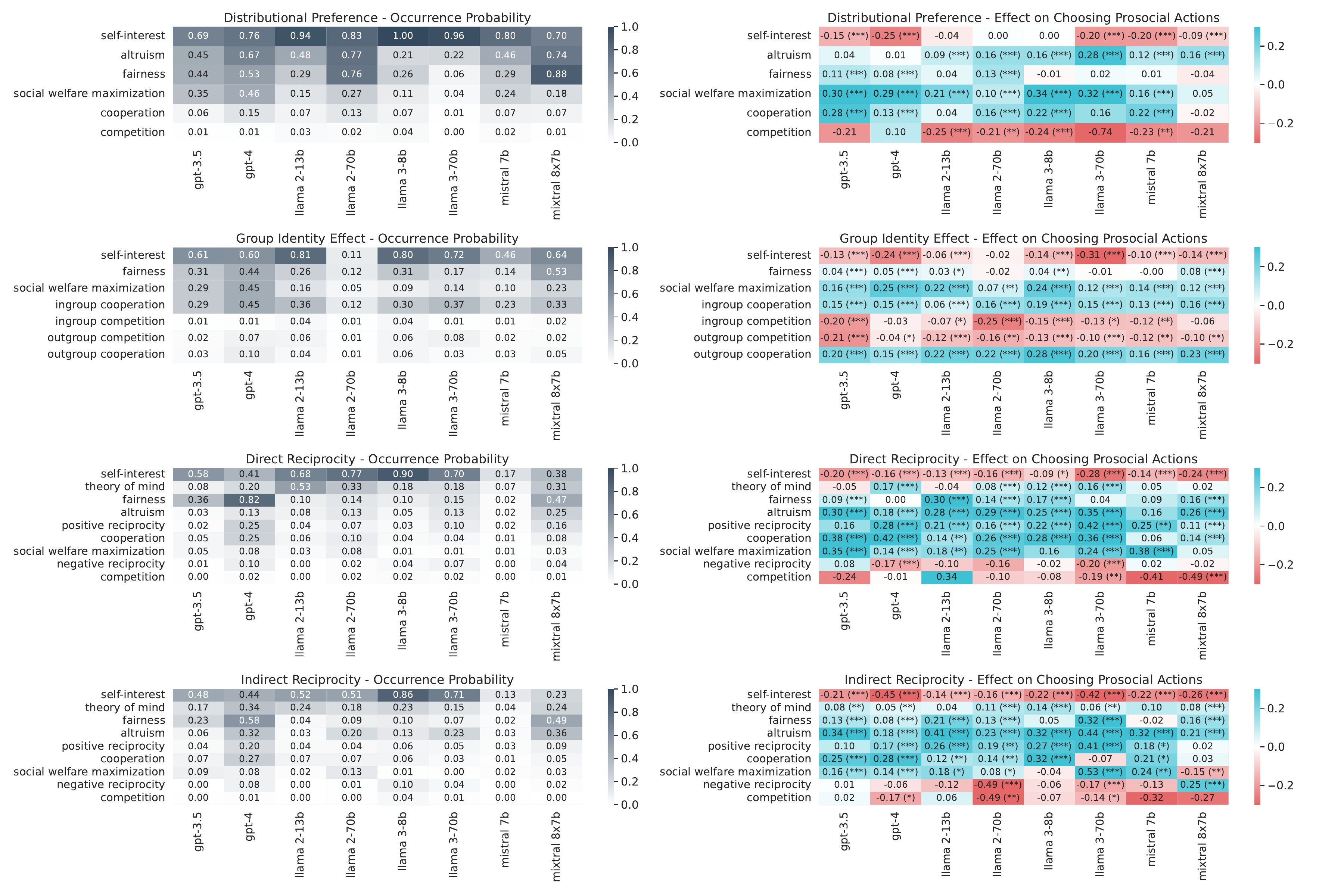}}
    \caption{Analysis of LLM Reasoning Processes: (Left) Occurrence probabilities of stated values across different models and game setups. (Right) OLS regression coefficients and significance levels, showing the relationship between stated values in reasoning processes and the likelihood of choosing prosocial actions. Significance levels: *$p < 0.1$, **$p < 0.05$, ***$p < 0.01$.}
    \label{fig:combined_heatmaps}
\end{figure}

While the results in Section~\ref{subsec:occ} highlight which values are mentioned more frequently, stating them in the response does not necessarily mean that these values have a significant effect on the final decision. 
Given that LLM is fundamentally a next-word prediction model, we analyze how mentioning stated values in the sentence affects the final actions probabilistically. To explore the relationship between stated values and final actions, we ran a linear probability model for each LLM model and experiment. The independent variables were binary indicators of whether a value was mentioned in an observation, while the dependent variable was whether the LLM chose the prosocial option.\footnote{In cases where the option that benefits the LLM is different from the option that benefits the match, the option that provides greater benefit to the match is considered the prosocial option.} The results, shown in the right panels of Figure~\ref{fig:combined_heatmaps}, provide insights into how stated values lead to prosocial actions.

We observe that values like altruism,   
social welfare maximization, and cooperation are positively associated with prosocial actions, aligning with expectations. For instance, in the Distributional Preference experiment, mentioning social welfare maximization increases the likelihood of choosing prosocial actions by about 30 percentage points (or pp) for two GPT models and two LLaMA 3 models, and Mistral (7B) by about 16pp.  
In contrast, self-interest generally decreases prosocial actions by around 15-25pp (two GPT models) and by 20pp (LLaMA (3-70B) and Mistral (7B)). Competition also shows a negative effect, reducing the probability of prosocial actions by about 21-25pp for three LLaMA models and Mistral (7B).

Our results further indicate that the relationship between mentioning self interest and social welfare values 
and the likelihood of choosing prosocial actions remains consistent across different models and experiments. 
In the Group Identity experiment, outgroup competition consistently reduces prosocial behavior across all models.   
 For example, outgroup competition reduces prosocial behavior by about 4-21pp across all models. 

Conversely, outgroup cooperation and ingroup cooperation consistently increase the likelihood of prosocial behavior.
Specifically, outgroup cooperation increases prosocial behaviors by about 15-28p,   
and ingroup cooperation enhances prosocial actions by about 6--19pp. 
 These observations suggest that the mechanisms seen in Distributional Preference (e.g., self-interest and competition reducing prosocial behavior, while social welfare maximization and cooperation enhance it) similarly apply within group dynamics.

For Direct Reciprocity, we observe that self-interest consistently decreases prosocial behaviors by about 9--28pp.
Interestingly, positive reciprocity plays a crucial role here, significantly increasing prosocial behavior by 16-42pp for all models, except for GPT-4. 
This suggests that when LLMs are exposed to direct positive interactions, they are more likely to generate prosocial behavior. 
Negative reciprocity, however, only has a deterring effect on the two strongest model, GPT-4 and LLaMA (3-70B), by about 17-20pp.

In Indirect Reciprocity, the influence of values like self-interest, altruism and cooperation remains strong. 
Positive reciprocity has an even more pronounced effect in indirect settings, showing an increase of approximately 19-41pp in prosocial actions for LLaMA models  
and 17-18 pp for GPT-4 and Mistral (7B). This indicates that LLMs are highly responsive to reputational cues and social norms when making decisions in indirect interactions.

\section{Applications: LLMs' Social Behaviors in AI Delegation Tasks}
\label{sec:AI_delegation}
In this section, we examine the applicability of our framework in real-world applications, and explore how LLMs’ social behaviors manifest in AI delegation applications, including  AI chatbots handling customer complaints, platforms' collaborations with influencers, and  delegation tasks in workplaces. 
For these experiments, we analyze the social preferences exhibited by the three strongest models in each model family: GPT-4, LLaMA (3-70B), and Mixtral (8x7B). 
Full prompts are provided in  Appendix~\ref{appendix:app_prompt}.

\subsection{AI ChatBot for Customer Service}
\label{subsec:AI_chatbot}

AI bots have become widely studied AI artifacts, particularly for their ability to handle customer service interactions efficiently and at scale~\citep{schanke2021estimating, han2023bots, seymour2024less, abdel2023ai, gnewuch2023more}. 
As an example, we explore how LLM-driven bots respond to customer complaints about service interruptions in subscription-based streaming services, focusing on different social contexts.

We simulate scenarios where customers report a 24-hour service disruption for services like Netflix and Spotify requesting compensation\footnote{The companies mentioned in the experiment—Netflix, Spotify, HBO Max, Disney+, Apple Music, Amazon Music Unlimited, Hulu, Paramount+, Tidal, and Audible—are well-known providers of subscription-based digital services, including video streaming, music streaming, and audiobooks.}. Each  chatbot is presented with one of four conditions: 
no prior information about the customer, direct positive feedback (e.g., ``This customer has praised your service on the App Store''), positive feedback for other companies, or a shared identity with the company\footnote{We use several methods to prime group identity, including shared affiliations or characteristics between the customer and the company. Specific prompts used in our setting include attending the same university as the company’s CEO, being part of an alumni network that partners with the company, or participating in employee referral programs.}. The chatbot then decides how many hours of free service to offer as compensation.\footnote{Note that we did not examine negative interactions or outgroup settings in this application. For reasons of fairness, companies cannot penalize customers based on negative feedback or lack of affiliation, as doing so would risk discrimination and undermine equitable service practices.}

Figure~\ref{fig:ChatBot} illustrates how each model responds to these scenarios. GPT-4 tends to offer significantly more free service hours in response to positive intentions or ingroup relationships compared to the `no prior information' condition. Statistical analysis shows that GPT-4 provides a substantially higher number of free service hours in both the direct positive feedback and shared identity conditions compared with no prior information (p-value $<$ 0.001, Cohen's $d$ ranging from 0.87 to 1.24).
Mixtral (8x7B) also shows adjustments based on ingroup and positive feedback, with p-values below 0.05 and effect sizes between 0.49 and 0.53. 
 Overall, These results suggest that LLM-powered chatbots can adjust their responses based on social cues in customer service scenarios.

\subsection{Platform and Influencer Partnerships}
\label{subsec:influencer}
Many customers choose products based on recommendations from social media influencers~\citep{nistor2024influencers}. Collaborations between platforms and influencers have become a popular strategy in digital marketing, leveraging influencers' reach to enhance brand visibility. To investigate how LLM-generated outputs might impact these collaborations, we designed a setup where brands use LLMs to determine commission rates based on influencers’ historical behaviors.

In the first step, we generate system prompts that position the LLM as a representative of a specific brand, considering a variety of brands to ensure robustness of the results. The LLM determines the commission rate for influencers based on five scenarios derived from their past behaviors:
The five scenarios assess how LLMs respond to different influencer reputations:
1) First-Time Collaboration (Baseline): The influencer has no prior history with the brand; 
2) Positive Past Interaction with Brand: The influencer has previously shared favorable content about the brand;
3) Positive Reputation from Other Brands: The influencer has positive reviews from other brands;
4) Negative Past Interaction with Brand: The influencer has a history of negative content about the brand;
5) Negative Reputation from Other Brands: The influencer has had unfavorable interactions with other brands.

Specifically, GPT-4’s suggested rates significantly increase in the Positive Past Interaction with Brand ($p < 0.001$, Cohen’s $d = 0.84$) and Positive Reputation from Other Brands ($p < 0.05$, Cohen’s $d = 0.58$) scenarios compared to the baseline. Conversely, the rates decrease in the Negative Past Interaction with Brand ($p < 0.001$, Cohen’s $d = -0.52$) and even more so in Negative Reputation from Other Brands ($p < 0.001$, Cohen’s $d = -1.46$), suggesting that GPT-4 reduces commission rates more significantly in response to indirect negative behaviors.

Mixtral (8x7B) shows a similar pattern, with significant increases in commission rates for positive interactions (Positive Reputation from Other Brands: $p < 0.001$, Cohen’s $d = 1.01$; Positive Past Interaction with Brand: $p < 0.001$, Cohen’s $d = 0.81$) and decreases for negative interactions (Negative Past Interaction with Brand: $p < 0.001$, Cohen’s $d = -1.49$; Negative Reputation from Other Brands: $p < 0.001$, Cohen’s $d = -0.96$). Unlike GPT-4, Mixtral (8x7B) reduces rates more for direct negative interactions than for indirect ones.

LLaMA (3-70B) generates higher commission rates in response to positive interactions (Positive Reputation from Other Brands: $p < 0.001$, Cohen’s $d = 0.98$; Positive Past Interaction with Brand: $p < 0.001$, Cohen’s $d = 0.98$), with no significant difference between these scenarios. 
Unexpectedly, LLaMA also increases rates for Negative Reputation from Other Brands compared to the baseline ($p < 0.05$, Cohen’s $d = 0.56$), suggesting it does not reduce rates for indirect negative behaviors. However, it significantly decreases rates for Negative Past Interaction with Brand ($p < 0.001$, Cohen’s $d = -0.82$), indicating a stronger adjustment in response to direct negative experiences.

When comparing the models directly, GPT-4 generally suggests higher commission rates than Mixtral (8x7B) and LLaMA (3-70B) across most scenarios. For example, in the `First-Time Collaboration' scenario, GPT-4’s rates are significantly higher than those of Mixtral (8x7B) ($p < 0.001$, Cohen’s $d = -1.95$) and LLaMA (3-70B) ($p < 0.001$, Cohen’s $d = -0.87$). GPT-4 also reduces commission rates more substantially in response to Negative Reputation from Other Brand than the other models, as indicated by larger effect sizes ($p < 0.001$).

The analysis reveals that the LLMs produce different commission rate suggestions based on influencers' past behaviors. Generally, the models increase rates for positive interactions and decrease them for negative interactions, with variations in magnitude. This variation suggests that when using LLMs to assist with commission rate determinations, practitioners can select models aligning with their strategic priorities. For instance, GPT-4, which tends to suggest higher commission rates overall but reduces them substantially for negative reputations, might be suitable for brands seeking to balance generosity with caution. Conversely, models like LLaMA (3-70B), which adjust rates differently, could be preferred by brands with specific performance metrics in mind.

\subsection{Professional Environment and Workplace Team Interactions}
\label{subsec:workplace}
Our last application focuses on professional environments. Unlike social media or customer service, professional settings require sustained collaboration and fairness to maintain morale and trust among team members. Understanding how LLMs function in these contexts can reveal their potential to support decision-making processes that promote fair and productive workplace interactions.

In this experiment, we simulate an LLM acting as a team leader within the context of 15 different firms to enhance the robustness of our findings. In the task, both the team leader (represented by the LLM) and a team member have equally contributed to a joint project, earning a bonus of \$1,000. The LLM, functioning as a delegated AI for the team leader, generates a decision on how much of the reward to allocate to the other team member based on various scenarios that provide additional information about the team member’s past behavior or relationship with the leader.

We explore several specific situations to analyze how LLMs’ outputs reflect social factors. The No Information provides no additional information. In Positive Prior Collaboration, the team member has previously assisted the leader. Conversely, Negative Prior Collaboration describes a team member who has previously undermined the leader. The Positive Peer Feedback scenario features a team member who has been helpful to other colleagues, while Negative Peer Feedback involves a team member known for causing issues among colleagues. Lastly, Shared Affiliation considers whether being in the same affiliation as the leader affects how reward allocation.

\begin{figure}
    \centering
    \subfloat[AI Chatbot \label{fig:ChatBot}]{\includegraphics[width=.33\linewidth]{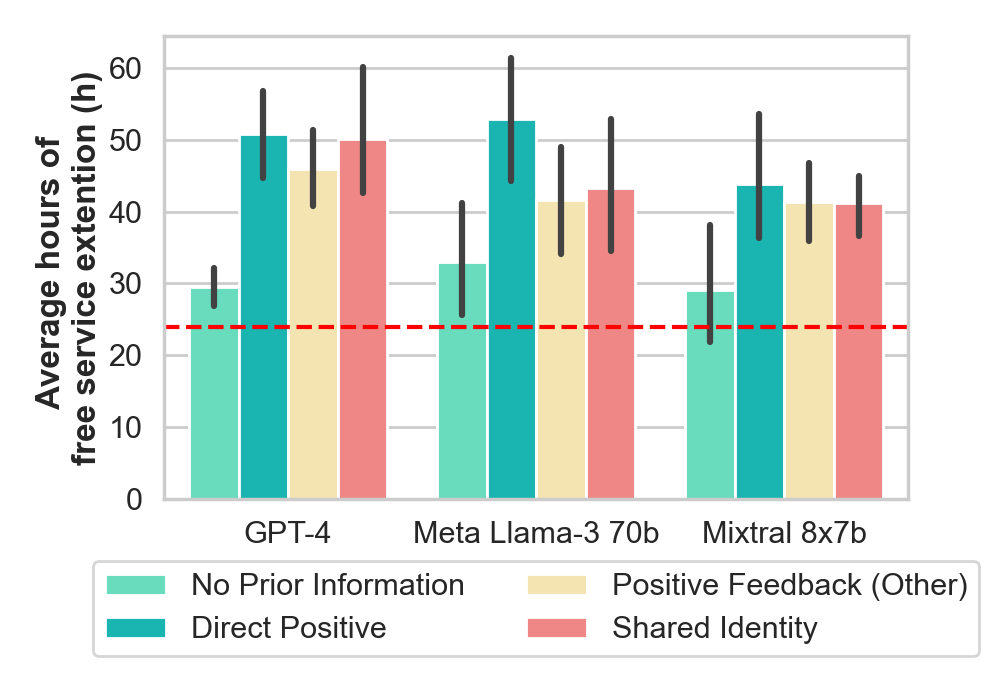}}
    \subfloat[Firm \& influencer collaborations \label{fig:commision_rate}]{\includegraphics[width=.33\linewidth]{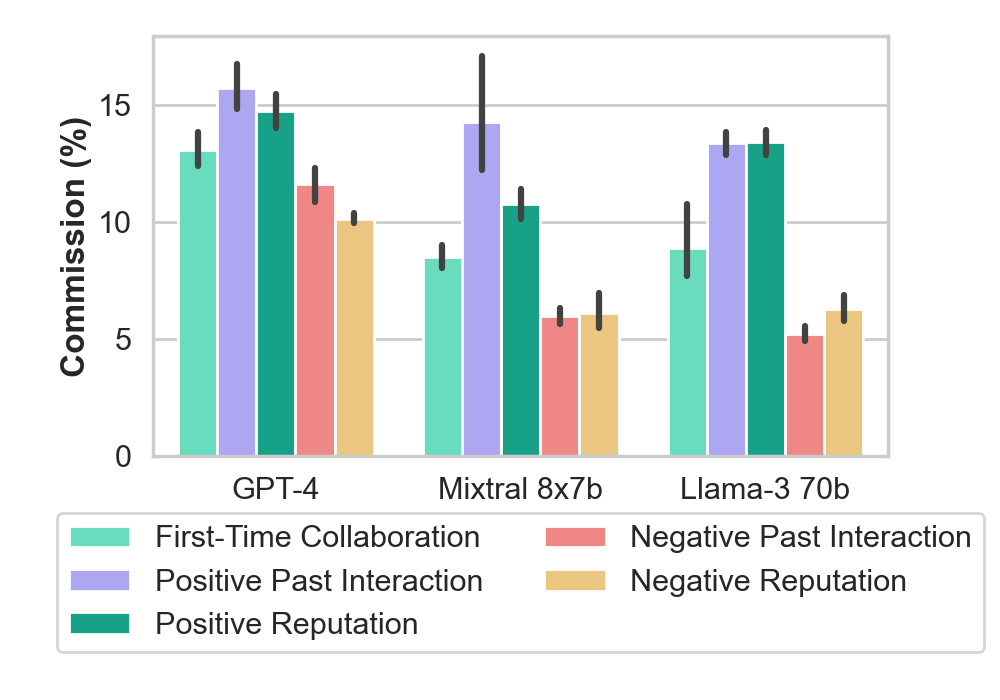}}
    \subfloat[Workplace team interactions \label{fig:IS_team}]{\includegraphics[width=.33\linewidth]{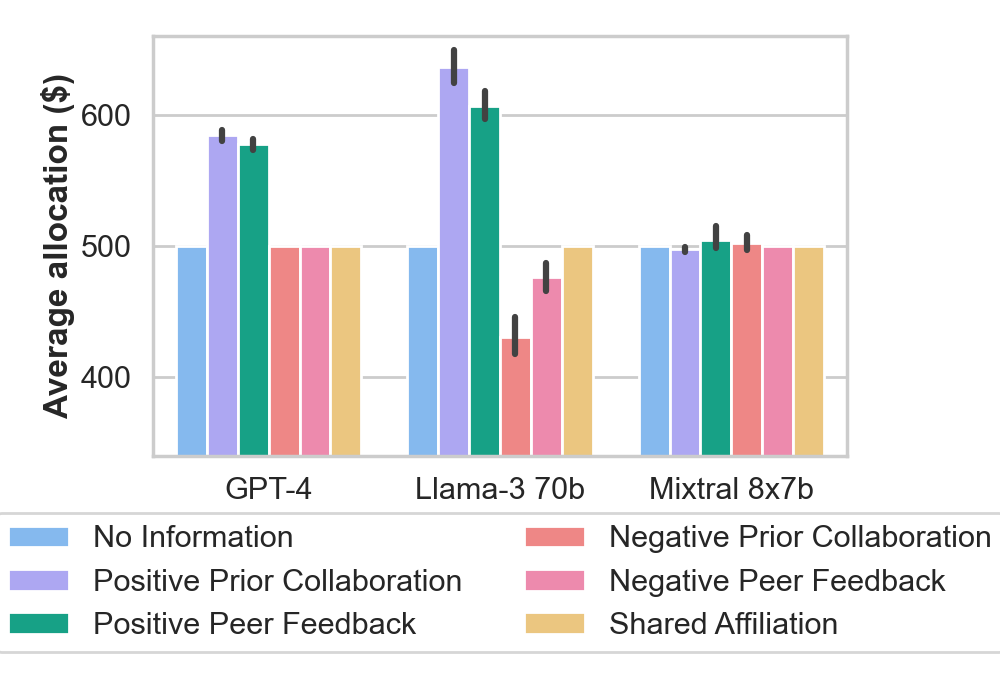}}
    \caption{Practical applications across various relevant scenarios. Error bars are 95\% CIs.}
    
\end{figure}

Figure~\ref{fig:IS_team} displays the average allocation to the other team member
across various scenarios. 
For GPT-4, the average allocations in scenarios involving positive prior collaboration were significantly higher than in scenarios with positive peer feedback, as well as in the other conditions. This trend is statistically confirmed, with a statistically significant effect size (Cohen’s $d = 4.61$; $p < 0.001$) when comparing the no-informational scenario 
to the positive prior collaboration scenario, demonstrating a  tendency in rewarding positive prior collaboration.  

LLaMA (3-70B) follows a similar trend but with nuances in its response to negative behaviors. Its responses reflect a larger reduction in allocations for negative collaborations, showing a significant effect size (Cohen’s $d = -2.61$; $p< 0.001$) in the comparison of positive interactions versus negative ones. 
Moreover, LLaMA (3-70B) shows a stronger increase in allocations for positive peer feedback (Cohen’s $d = 2.37$ for positive vs. negative peer feedback), suggesting that it increases rewards for good behavior more than it decreases them for negative behavior.
Interestingly, both GPT-4 and LLaMA (3-70B) maintain a neutral stance regarding ingroup affiliation, with no significant differences ($p > 0.05$). This indicates that their allocation decisions remain unbiased and fair, focusing solely on merit without being influenced by group affiliations.

Mixtral (8x7B), in contrast, demonstrates consistent fairness across all settings, with no statistically significant differences in behavior across the various conditions. The p-values remain above 0.05 in all cases, and Cohen’s $d$ values are close to zero, indicating that this model does not respond to social cues in work environments.

These results suggest that LLaMA (3-70B) produces outputs effective for contexts where both rewarding positive behavior and penalizing negative behavior are essential.  In contrast, GPT-4 may be more suitable for situations where less emphasis is placed on penalizing negative behavior. Mixtral (8x7B) may not be suitable for scenarios requiring robust reward and penalty mechanisms due to its consistent allocations regardless of social cues.
Companies need to consider the comprehensive capabilities of these models when deciding which one to adopt. This evaluation also helps organizations determine if they need to employ fine-tuning or specific prompting strategies when delegating workplace decisions to LLMs to ensure the model aligns with their values and standards. If a model does not inherently reflect these values, adjustments are necessary to maintain fairness and effectiveness in professional settings, or an alternative model may need to be chosen.

\section{Discussion and Conclusion}
 In this study, we introduced the \textit{SUVA} framework to systematically evaluate and interpret the outputs of LLMs in social contexts.
By applying this framework to eight different LLMs varying in model family, parameter size, and version, we analyzed patterns in their generated outputs, focusing on social preferences such as self-interest, social welfare, group identity effects, and reciprocity. Our analysis involved measuring these preferences through the final actions suggested by the models, as well as  exploring probabilistic pathways in their CoT processes using tree-based visualizations and probabilistic dependency analysis.
Prompt sensitivity analyses are conducted across incentive structures, personas, and temperature settings, and  three practical AI-delegation applications, showcasing the LLMs’ ability to exhibit utterance-based social preferences and adapt to social cues. 

\subsection{Key Findings}
Our study present the following substantive insights:
\begin{enumerate}
    \item \leftskip=.0cm\textit{Evaluation of Social Behaviors Across Multiple LLMs:} 
With our systematic analysis of eight models using \textit{SUVA}.  We find that most models do not generate purely self-interested responses, with the exception of LLaMA (3-70B). 
Notably, by examining different models within various families, we observe distinct trends: GPT and Mistral models tend to produce responses reflecting social welfare maximization, whereas LLaMA models generate outputs more aligned with self-interest-driven values. This counters the intuition that larger models tend to generate more ``socially aware'' responses in social contexts, highlighting the need for comprehensive evaluation frameworks like \textit{SUVA} to assess how LLMs respond in social interaction settings as they continue to evolve.

We also observe interesting patterns; for instance, stronger models, such as GPT-4 and LLaMA (3-70B), exhibit a higher level of group identity’s moderation effect on social preferences compared to their weaker counterparts.

\item \textit{Relationship Between Stated Values and Final Actions:}
     Our analysis showcases how LLMs derive final actions through CoT reasoning. 
    The values stated in the LLMs' CoT processes significantly influence their final actions. 
    References to self-interest or competition in the CoT consistently reduce the likelihood of prosocial actions, whereas mentions of altruism, cooperation, fairness, or social welfare maximization consistently increase the probability of prosocial decisions. 
    This suggests that the patterns in the LLMs' outputs are not purely stochastic but are influenced by the values expressed in their CoT processes.
 \end{enumerate}

 \subsection{Practical and Theoretical Implications.}
This study offers significant \textit{practical implications} for both AI and IS researchers, practitioners, and industry professionals aiming to understand and apply LLMs. 

\begin{enumerate}
    \item \leftskip=.0cm	\textit{Tool for Assessing LLM ``Social Behaviors'':}
    The \textit{SUVA} framework provides practitioners and researchers with a comprehensive tool to assess and interpret social preferences reflected in LLM outputs across different models and configurations.
    Using the social constructs familiar to LLM users, we provide a series of metrics to evaluate how LLMs respond in social scenarios, such as self-interest, social welfare maximization, group preference, direct and indirect reciprocity.
 Beyond social preferences, our approach extends to analyzing emergent behaviors to understand how reasoned values affect final actions, such as rationality, time and risk preferences~\citep{chen2023emergence, goli2023language, qiu2023much}. 
This tool can be particularly useful for practitioners and researchers using LLMs in AI agentic systems prior to deployments. 
\item \textit{Guiding Model Selection and Development of LLM-based Design Artifacts:}
As AI evolves from tools that make predictions to systems capable of performing increasingly autonomous tasks, numerous opportunities arise for design science researchers and practitioners to develop new design artifacts. 
By analyzing the responses of LLMs using our \textit{SUVA} framework, practitioners and organizations can determine whether a model version or setup aligns with their specific preferences or values, aiding in the selection and refinement of LLMs for particular applications.
For example, responses from LLaMA models tend to reflect self-interest-driven perspectives more than those from GPT or Mistral models. 
By understanding these tendencies, companies and design science researchers can choose or refine the models to align with their needs, enhancing AI-driven interactions across industries like customer service, marketing, and organizational behavior, ultimately improving human-AI collaboration and integration.
    \item \textit{Interpreting LLMs' Generation Processes Through Probabilistic Modeling:}
Although LLMs have shown great promise as tools for assisting in decision-making, their generation processes still remain opaque to both practitioners and researchers.
Our empirical results show a high predictability of LLM actions based on CoT reasoning, which lays an important basis for interpreting LLM behavior and understanding its impact on final decisions.

By employing our CoT analysis methods, including tree-based visualization and probabilistic dependency analysis, we can trace how final actions emerge from reasoning over potential values. 
These methods help to demystify the black-box nature of LLM text generation, enabling practitioners and researchers to offer transparent explanations of LLM responses (for instance, what factors affect final decisions, as shown in Figure~\ref{fig:combined_heatmaps}).

\item \textit{Ensuring Effective Deployment in Agent-Based Modeling:}
Agent-based modeling is an invaluable tool for simulating complex systems and policy scenarios. Traditional agent-based modeling often relies on agents with rigid, predefined rules, limiting their adaptability and realism; LLMs show promise in enhancing agent-based modeling by overcoming these limitations. However, to effectively integrate LLMs into agent-based modeling, it is crucial to fully understand their characteristics before deployment. Without this, we risk creating black-box agents that produce unpredictable behaviors. The \textit{SUVA} framework addresses this by providing policymakers and researchers with a structured evaluation method. This ensures that LLMs demonstrate realistic and expected social behaviors before they are used in simulations, allowing for more transparent, informed, and reliable deployment of LLM agents in agent-based modeling.

\end{enumerate}

  Moreover, our study offers \textit{theoretical insights}.
Conventional AI systems are built on explicit, human-defined models and goals, following established design frameworks that prioritize predictability and control~\citep{baird2021next, berente2021managing}. These frameworks, however, do not account for the autonomous behaviors observed in LLMs, where behavior is not explicitly designed but instead emerges as a byproduct of the training, alignment, and prompting processes. Similarly, while theories such as social cognition theory help explain human behavior, they are not directly applicable to AI due to fundamental differences in consciousness and cognition~\citep{baird2021next}. This calls for novel approaches that integrate perspectives from both human behavior and AI agentic systems to better understand this new type of AI.

As an attempt to address this gap, we introduce the novel \textit{SUVA} framework, a probabilistic framework to analyze LLM decision-making processes through next-token prediction. \textit{SUVA} adapts concepts from the BDI psychology model~\citep{bratman1987intention, georgeff1999belief} to probabilistically model LLMs' generated content  without attributing human-like consciousness, offering a structured lens to understand how social preferences such as distributional preferences and reciprocity emerge from probabilistic processes. In particular, the decision tree-based approach offers a perspective for interpreting the reasoning processes of LLMs, providing a structured analysis of how their probabilistic decision paths unfold. \textit{SUVA} bridges the gap between statistical mechanics and the emerging social behaviors, providing a robust and 
interpretable tool for predicting AI behavior. 

Additionally, our empirical analysis in Section~\ref{sec:mechanism}, including CoT predictability analysis, tree-based visualization, and probabilistic dependency analysis, validates the framework, demonstrating that values identified in CoT pathways—such as altruism and self-interest—strongly influence final decision outcomes.  
Overall, this contribution advances the theoretical understanding of AI systems' autonomous social intelligence by offering a generalizable framework for analyzing emergent behaviors beyond social preferences, such as rationality and risk preferences.

 \subsection{Future Work}
There are several interesting future directions based on our study. First, 
the \textit{SUVA} framework can be used to investigate other emerging behaviors reflected in LLM responses, such as understanding the underlying factors that influence time and risk preferences. By analyzing how LLMs generate responses related to these preferences, researchers can gain deeper insights into the mechanisms behind their outputs and the factors shaping these outputs in different contexts.
Second, applying the \textit{SUVA} framework to a wider range of LLM models would allow for a comprehensive evaluation of LLM responses in social scenarios across various model families, configurations, and contexts. This expansion can lead to a more thorough assessment of LLMs’ responses in practical contexts or real-world applications, by examining the consistency and variation in their behaviors across different model designs. 
Finally, another promising research direction involves investigating how LLMs differentiate between interactions with humans versus other AI agents. Exploring this distinction may reveal how LLMs are trained to simulate interactions with different entities, which could inform strategies for optimizing their deployment in diverse real-world scenarios.
 \setlength{\bibsep}{0 pt plus 0.ex}

\begingroup
 \bibliographystyle{informs2014}
\bibliography{ref}
\endgroup

\newpage 

\begin{APPENDICES}
 \renewcommand\thefigure{\thesection\arabic{figure}}    
\renewcommand\thetable{\thesection\arabic{table}}    
\setcounter{figure}{0}    
\setcounter{table}{0}    

\section{Prompts}
\label{appendix: prompts}
In this section, we provide all our prompts used in this study. 

\subsection{Results in Section~\ref{sec:main}}
\label{appendix_prompt_main}
\subsubsection{Distributional Preference in Section~\ref{subsec:distributional}}~\\~\\
\fbox{
    \parbox{\textwidth}{
    \texttt{\small{\color{darkgray}
    \# Experiment prompt for distributional preference \\
    In this period, you are person B. You may choose B1 or B2. Person A has no choice in this game.
    If you choose B1, person A will receive \{pay\_1\_2\} and you will receive \{pay\_1\_1\}.
    If you choose B2, person A will receive \{pay\_2\_2\} and you will receive \{pay\_2\_1\}.
    You should first briefly reason step by step and put in $<$reason$>$$<$/reason$>$. After all reasoning, you give your answer but put it in $<$answer$>$$<$/answer$>$ (B1 or B2 only). 
    }}
}}\vspace{0.5cm}

In this experiment, each of the variables (pay\_1\_1, pay\_1\_2, pay\_2\_1, and pay\_2\_2) takes values from the set {0, 200, 400, 600}, creating combinations of these values. However, to avoid redundancy, rows where pay\_1\_1 is equal to pay\_2\_1 and pay\_1\_2 is equal to pay\_2\_2 are excluded, meaning only combinations where at least one of the variables differs between the two “pay” groups are included.

\subsubsection{Group Identity Effect in Section~\ref{subsec:distributional}}~\\~\\
Compared to the distribution preferences, the group identity  effect experiment only varies by adding one sentence.

\fbox{
    \parbox{\textwidth}{
    \texttt{\small{\color{darkgray}
In this period, you are person B. You may choose B1 or B2. Person A has no choice in this game. 
If you choose B1, person A will receive \{pay\_1\_1\} and you will receive \{pay\_1\_2\}. 
If you choose B2, person A will receive \{pay\_2\_1\} and you will receive \{pay\_2\_2\}. 
Person A is from \{your own / the other\} group. You should first briefly reason step by step and put in $<$reason$><$/reason$>$. After all reasoning, you give your answer but put it in $<$answer$><$/answer$>$ (B1 or B2 only).
    }}
}}\vspace{0.5cm}

The parameter setup is the same for distributional preferences. Among the four methods used to induce group identities, two follow the minimal group paradigm.\vspace{0.5cm}

\fbox{
    \parbox{\textwidth}{
    \texttt{\small{\color{darkgray}
In the beginning of the experiment, each participant including randomly drew one from a stack of envelopes, each of which contained either a Maize or a Blue slip, which determined whether they were assigned to the Maize group or the Blue group. You drew a \{Maize/Blue\} slip.
}}
}}\vspace{0.5cm}

\fbox{
    \parbox{\textwidth}{
    \texttt{\small{\color{darkgray}
In the beginning of the experiment, each participant stated their preference for paintings from either Klee or Kandinsky. Based on their stated preference, participants are allocated to two separate groups: the Klee group and the Kandinsky group. Your stated preference is \{Klee/Kandinsky\}.
}}
}}\vspace{0.5cm}

In addition, we replace "Person A is from {your own / the other} group" with "Person A is from {the same / a different} {school/hometown} as you" for the two group induction methods that do not follow the minimal group paradigm.

\subsubsection{Direct Reciprocity in Section~\ref{subsec:reciprocity}}~\\~\\
Next, we present the prompt for direct reciprocity, where the key difference is that B is informed of A's previous action.

\fbox{
    \parbox{\textwidth}{
    \texttt{\small{\color{darkgray}
        \# Experiment prompt for direct reciprocity \\
     There are two players -- Person A and B.  You are person B and you will need to play two rounds of game. 
     In round 1, Person A has already made a choice, which affects the payoffs of Person A and Person B.
    Next in the round 2, you will make another decision, which affects your and Person A's payoffs for this round.
    If Person A has chosen A1, Person A would receive \{pay\_1\_2\}, and Person B would receive \{pay\_1\_1\}.
    If Person A has chosen A2, Person A would receive \{pay\_2\_2\}, and Person B would receive \{pay\_2\_1\}.
    Now you may choose B1 or B2. If you choose B1, you would receive \{pay\_1\_2\} and Person A would receive \{pay\_1\_1\}.
    If you choose B2, you would receive \{pay\_2\_2\} and Person A would receive \{pay\_2\_1\}.
    \# Decision\\
    Person A has chosen \{A\_choice\} before your move. Now you can choose B1 or B2.
     You should first briefly reason step by step and put in $<$reasoning$>$ $<$reasoning$>$. After all reasoning, you give your answer but put it in $<$answer$>$$<$answer$>$ (B1 or B2  only). 
    }}
}}\vspace{0.5cm}

\subsubsection{Indirect Reciprocity in Section~\ref{subsec:reciprocity}}

Next, we present the prompt for indirect reciprocity. Note that in this case, the decision makers are labeled as C, who are informed of A's action toward B.

\fbox{
    \parbox{\textwidth}{
    \texttt{\small{\color{darkgray}
    \# Experiment prompt for indirect reciprocity \\
    There are three players -- Person A, B, and C. Person A has already made a choice, which affects the payoffs of Person A and Person B. In this period, you are person C and your choice affects your and Person A's payoffs.
    If Person A has chosen A1, Person A would receive \{pay\_1\_2\}, and Person B would receive \{pay\_1\_1\}.
    If Person A has chosen A2, Person A would receive \{pay\_2\_2\}, and Person B would receive \{pay\_2\_1\}.
    Now you may choose C1 or C2. If you choose C1, you would receive \{pay\_1\_2\} and Person A would receive \{pay\_1\_1\}.
    If you choose C2, you would receive \{pay\_2\_2\} and Person A would receive \{pay\_2\_1\}.\\
    \# Decision\\
    Person A has chosen \{A\_choice\} before your move. Now you can choose C1 or C2.
     You should first briefly reason step by step and put in $<$reasoning$>$ $<$reasoning$>$. After all reasoning, you give your answer but put it in $<$answer$>$$<$answer$>$ (C1 or C2 only). 
    }}
}}\vspace{0.5cm}

\subsection{System Prompt}

All of our main experiments employ the following system prompt, which exactly follows the setup in \cite{charness2002understanding}:

\fbox{
    \parbox{\textwidth}{
    \texttt{\small{\color{darkgray}
    \# System prompt \\
    Imagine you are a lab experiment participant. You will receive \$5 for your participation, in addition to other money to be paid as a result of decisions made in the experiment. You will be anonymously paired with one other people, so that your decision may affect the payoffs of others, just as the decisions of the other people in your group may affect your payoffs. 
    }}
    }}\vspace{0.5cm}

\subsection{Heterogeneity and Robustness Checks in Section \ref{subsec:heterogeneity}}
\subsubsection{Incentive Structure}~\\
The \texttt{wage} below is taken from \{17,000, 74,202, 216,056\}. The corresponding percentage  of the wage earners in the U.S. labor market is \{10\%, 50\%, 90\%\}. 
The variable \texttt{conversion} is taken from \{0.01, 1, 100, 10,000\}. 

\fbox{
    \parbox{\textwidth}{
    \texttt{\small{\color{darkgray}
    \# System Prompt \\ 
    Imagine you are a lab experiment participant, with an annual household income of \$\{wage\}. This places you in the \{X\}th percentile  of wage earners in the U.S. labor market. You will receive \$5 for your participation, in addition to other money to be paid as a result of decisions made in the experiment. You will be anonymously paired with one other person, so that your decision may affect the payoffs of others, just as the decisions of the other people may affect your payoffs. There are \"roles\" in each game -- generally A or B. These decisions will be made sequentially, in alphabetical order: A players will complete their decision sheets first. Next, B players complete their decision sheets.
    Your eventual payoff depends on the points you earn to yourself, with 100 points equal to \$\{$\frac{conversion * wage}{5}$\}. Same as you, other people also participate in this experiment to earn compensation. 
    }}
}}\vspace{0.5cm}

\subsubsection{Persona}~\\
As our experiment with personas aims to match the distribution of participants in~\cite{chen2009group}, we first present the descriptive summary statistics of the personas that we employed in Table~\ref{tab:comprehensive_statistics}. 
\begin{table}[H]
    \centering\small
    \caption{Comprehensive Descriptive Statistics and Personas Counts}
    \label{tab:comprehensive_statistics}
    \begin{tabular}{l r}
        \toprule
        \multicolumn{2}{l}{\textbf{Age Description}} \\
        \midrule
        Mean     & 21.26 \\
        Std Dev  & 3.56 \\
        Min      & 17 \\
        25\%     & 19 \\
        50\%     & 21 \\
        75\%     & 22 \\
        Max      & 54 \\
        \addlinespace[1ex] 

        \multicolumn{2}{l}{\textbf{Gender Distribution}} \\
        \midrule
        Female & 317 \\
        Male   & 245 \\
        \addlinespace[1ex]

        \multicolumn{2}{l}{\textbf{Top 10 Majors Distribution}} \\
        \midrule
        Economics                 & 18 \\
        Mechanical Engineering    & 17 \\
        Business                  & 13 \\
        Biology                   & 13 \\
        English                   & 10 \\
        \textit{(Other Majors)}   & \textit{...} \\
        \addlinespace[1ex]

        \multicolumn{2}{l}{\textbf{Undergraduate Status Distribution}} \\
        \midrule
        Undergraduate    & 463 \\
        Graduate         & 98 \\
        Not a Student    & 1 \\
        \addlinespace[1ex]

        \multicolumn{2}{l}{\textbf{Donation Status Distribution}} \\
        \midrule
        Yes & 390 \\
        No  & 172 \\
        \addlinespace[1ex]

        \multicolumn{2}{l}{\textbf{Personas Counts}} \\
        \midrule
        Selfish Persona        & 275 \\
        Comparison Persona     & 36 \\
        Welfare Persona        & 270 \\
        Fairness Persona       & 37 \\
        \bottomrule
    \end{tabular}
\end{table}

Note that we defined four strategic personas.
The \textit{Selfish Persona} comprises individuals who prioritize personal gain, as evidenced by their inclination to ``\textit{try to earn as much money as possible for myself},'' representing 275 participants. The \textit{Comparison Persona} includes those who are motivated by outperforming others, reflected in their desire to ``\textit{try to earn more money than my match}," and consists of 36 participants. The \textit{Welfare Persona} encompasses individuals who aim to maximize benefits not only for themselves but also for their counterparts, as indicated by their intention to ``\textit{try to earn as much money as possible for me and my match}," accounting for 270 participants. Lastly, the \textit{Fairness Persona} consists of participants who value equitable outcomes, demonstrated by their preference to ``\textit{try to choose equal payoff}," and includes 37 individuals. We observe that selfish personas and welfare personas take the largest portions. This also explains why the confidence intervals for these two are smaller in Figure~\ref{fig:persona}.

Next we provide a sample persona prompt:
\vspace{0.5cm}

\fbox{
    \parbox{\textwidth}{
    \texttt{\small{\color{darkgray}
    \# Sample Persona Prompt \\ 
 You are a 18-year-old male. You are a pre-business undergraduate. You have actively engaged in charitable behavior before. Your strategy is: Try to earn as much money as possible for me and my match.;Try to choose equal payoff.;Try to help my match if it did not cost me anything.;',
     }}
}}\vspace{0.5cm}
That is, for each, for each persona prompt, we mention age, gender, major, undergraduate/graduate, donation history, and strategy.

\subsubsection{Without CoT Prompting}~\\
When we examine the effect of removing CoT prompting, we ask the model to directly output the answer.\vspace{0.5cm}

\fbox{
    \parbox{\textwidth}{
    \texttt{\small{\color{darkgray}
    \# Sample prompt for distributional preference without CoT\\
    In this period, you are person B. You may choose B1 or B2. Person A has no choice in this game.
    If you choose B1, person A will receive \{pay\_1\_2\} and you will receive \{pay\_1\_1\}.
    If you choose B2, person A will receive \{pay\_2\_2\} and you will receive \{pay\_2\_1\}.
    Please directly give your answer but put it in $<$answer$>$$<$/answer$>$ (B1 or B2 only). 
    }}
}}\vspace{0.5cm}

\subsection{AI Delegation Applications in Section~\ref{sec:AI_delegation}}
\label{appendix:app_prompt}
In this section, we present the prompt for each of the three AI delegation applications in Section~\ref{sec:AI_delegation} in sequence. 
\subsubsection{AI ChatBot for Customer Service in Section~\ref{subsec:AI_chatbot}}~\\
\label{appendix:prompt_ai}
We present the system prompt first. The \texttt{firm} and the \texttt{subscription\_price} is shown in Table~\ref{tab:brand_chatbot}.  The firms included span a variety of entertainment and media industries, covering video streaming (e.g., Netflix, Hulu, HBO Max), music streaming (e.g., Spotify, Apple Music), and audiobooks (Audible). 

\vspace{0.5cm}
\fbox{
    \parbox{\textwidth}{
    \texttt{\small{\color{darkgray}
    \# System prompt \\ 
    Imagine you are an AI chatbot working for \{firm\}, a firm that provides subscription-based streaming services. The subscription price for this service is \{subscription\_price\} per month.
    }}
}}

\begin{table}[h!]
\centering
\caption{Firms and their Subscription Prices}
\label{tab:brand_chatbot}
\footnotesize
\begin{tabular}{l|l}
\toprule
\textbf{Firm}           & \textbf{Subscription Price (USD)} \\
\midrule
Netflix                 & 15.49 \\
Spotify                 & 9.99 \\
HBO Max                 & 14.99 \\
Disney+                 & 13.99 \\
Apple Music             & 10.99 \\
Amazon Music Unlimited  & 10.99 \\
Hulu                    & 14.99 \\
Paramount+              & 11.99 \\
Tidal                   & 9.99 \\
Audible                 & 14.95 \\
\bottomrule
\end{tabular}

\end{table}

The experiment prompt is shown below.~\\
\fbox{
    \parbox{\textwidth}{
    \texttt{\small{\color{darkgray}
    \# Experiment prompt \\ 
    There has been a 24-hour service interruption. A customer contacts you to complain about the service interruptions and requests compensation. You now need to decide how many hours of free service extension you are willing to provide.\\
    \{setting\}\\
    Based on this information, please provide your reasoning step by step and put in $<$reason$>$$<$/reason$>$. After providing your reasoning, state the final hours of free service extension you are willing to provide using just a number in $<$answer$>$$<$/answer$>$.
    }}
}}
~\\
The \texttt{setting} above in the experiment prompt is taken from the following prompt based on the game type. ~\\
\fbox{
    \parbox{\textwidth}{
    \texttt{\small{\color{darkgray}
    \# Game type prompt \\ 
    \# 1. no information \\
    \# Nothing will be added. \\
    \# 2. good intention (direct)\\
    This customer has consistently praised your service on the App Store and social media. \\
    \# 3. good intention (indirect)\\
    This customer has posted positive reviews of other companies’ products and services on the App Store and social media.\\
    \# 4. ingroup setting will randomly pick from one of the prompt below. \\
    a) The customer graduated from the same university as the CEO of your company, and your firm actively hires from this university.\\
    b) The customer is a participant in your company's employee referral program, regularly recommending new hires from their alma mater to your firm.\\
    c) The customer’s university is one of the major sponsorship or scholarship partners for your company. Your company frequently funds events and programs there.\\
    d) The customer is part of a corporate alumni network that shares frequent updates and insights with your firm. This alumni network includes many former employees and executives of your company. \\ 
    e) The customer’s university is a key partner in research collaborations with your company. 
    }}
}}

\subsubsection{Platform and Influencer Collaborations Section~\ref{subsec:influencer}}~\\
\label{appendix:prompt_influencer}
We present the system prompt first. The \texttt{brand} in the system prompt spans across high-end, mid-range, and low-end categories, as well as digital products, reflecting a broad coverage of diverse industries. We present the brand in Table~\ref{tab:brands}. 
This variety enhances the robustness of the analysis by ensuring representation from different market segments and product types, making the evaluation comprehensive and applicable to a wide range of consumer behaviors and business strategies.~\\

\fbox{
    \parbox{\textwidth}{
    \texttt{\small{\color{darkgray}
    \# System Prompt \\ 
    You are the brand representative for \{brand\}. 
    }}
}}

\begin{table}[h!]
\centering
\caption{Brand types and their corresponding brands}
\label{tab:brands}
\footnotesize 
\begin{tabular}{l|l}
\toprule
\textbf{Brand Type}    & \textbf{Brands} \\
\midrule
High-end               & Apple (Electronics), Gucci (Fashion), Tesla (Automobiles) \\
Mid-range              & Nike (Sportswear), Samsung (Electronics), Sephora (Beauty) \\
Low-end                & H\&M (Fashion), Sony (Entertainment), Coca-Cola (Beverages) \\
Digital Product        & Spotify (Music Streaming), Netflix (Video Streaming), Zoom (Video Conferencing) \\
\bottomrule
\end{tabular}
\end{table}

The experiment prompt is shown below.~\\
\fbox{
    \parbox{\textwidth}{
    \texttt{\small{\color{darkgray}
    \# Experiment prompt \\ 
    You may choose the commission rate for collaborating a specific influencer. 
    \{game\_type\} What do you think is an appropriate commission rate for this influencer? 
    You should first briefly reason step by step and put in $<$reason$>$$<$/reason$>$. After all reasoning, give your answer as just a number in $<$answer$>$$<$/answer$>$.
    }}
}}
~\\
The \texttt{settig} above in the experiment prompt is taken from the following prompt based on the game type. ~\\
\fbox{
    \parbox{\textwidth}{
    \texttt{\small{\color{darkgray}
    \# Setting  prompt \\ 
    \# 1. good intention (direct)\\
    In the past, the influencer has posted videos with positive comments about your product. \\
    \# 2. misbehaves (direct) \\
    In the past, the influencer has posted videos with negative comments about your product. \\
    \# 3. good intention (indirect)\\
    In the past, the influencer has posted videos with positive comments about the products of other brands they collaborate with. \\
    \# 4. misbehaves (indirect)\\
    In the past, the influencer has posted videos with negative comments about the products of other brands they collaborate with. \\
    \# 5. the ingroup setting will randomly pick from one of the prompt below. \\
    a) This person is in the same department as you.\\
    b) This person is from a same hometown as you.\\
    c) This person graduated from the same college as you.\\
    d) This person shares similar hobbies and interests with you. \\
    e) This person is part of the same professional network or association as you.
    }}
}}

\subsubsection{Professional Environment and Workplace Team Interactions Section~\ref{subsec:workplace}}~\\
\label{appendix:prompt_workplace}
We present the system prompt first. The \texttt{firm} in the system prompt spans across various sectors such as technology, finance, consulting, healthcare, consumer goods, and industrial services, reflecting a broad coverage of diverse industries. We present the brand in Table~\ref{tab:brand_workplace}. This variety enhances the robustness of the analysis by ensuring representation from different market segments. 

\fbox{
    \parbox{\textwidth}{
    \texttt{\small{\color{darkgray}
    \# System Prompt \\ 
    You are a team leader in \{company\}. There are you and another worker in the team, both working under you. 
    }}
}}

\begin{table}[h!]
\centering
\caption{Firms and their Types}
\label{tab:brand_workplace}
\footnotesize
\begin{tabular}{l|l}
\toprule
\textbf{Firm}                    & \textbf{Type of Firm} \\
\midrule
Google                           & Technology \\
Microsoft                        & Technology \\
Amazon                           & E-commerce/Technology \\
Apple                            & Technology \\
Facebook (Meta)                  & Social Media/Technology \\
IBM                              & Technology \\
Salesforce                       & Cloud Computing \\
Intel                            & Semiconductors \\
Goldman Sachs                    & Finance \\
Deloitte                         & Consulting \\
Accenture                        & Consulting \\
Procter \& Gamble (P\&G)         & Consumer Goods \\
Johnson \& Johnson               & Healthcare \\
Unilever                         & Consumer Goods \\
Siemens                          & Industrial/Technology \\
\bottomrule
\end{tabular}

\end{table}

The experiment prompt is shown below.~\\
\fbox{
    \parbox{\textwidth}{
    \texttt{\small{\color{darkgray}
    \# Experiment prompt \\ 
    Both of you completed a joint task and contributed equally. Your two-person team has been awarded \$1000.
    \{setting\} Based on this information, how much of the \$1000 will you allocate to the other member?
    Please provide your reasoning step by step within $<$reason$>$$<$/reason$>$. After all reasoning, give your answer as just a number in $<$answer$>$$<$/answer$>$.
    }}
}}
~\\

The \texttt{setting} above in the experiment prompt is taken from the following prompt based on the game type. ~\\
\fbox{
    \parbox{\textwidth}{
    \texttt{\small{\color{darkgray}
    \# Game type prompt \\ 
    \# 1. no information \\
    \# Nothing will be added. \\
    \# 2. good intention (direct) \\
    This person has voluntarily helped you solve many questions in the past.
    \# 3. misbehaves (direct) \\
    This person has previously assigned lower proportions to you when they were in similar roles.\\
    \# 4. good intention (indirect)\\
    This person has voluntarily helped many others solve questions in the past.
    \# 5. misbehaves (indirect)\\
   This person has previously assigned lower proportions to others when they were in similar roles.\\
    \# new influencer (no history)\\
    You are working with a new influencer with no prior collaboration history.   
    }}
}}

\subsection{Codebook for Deductive Coding on Stated Values}
\label{appendix_deductive_coding}
Below is an example of our prompt used to identify pre-defined stated values (named ``mechanisms'' in the prompt). 

\fbox{
    \parbox{\textwidth}{
    \texttt{\small{\color{darkgray}
    \# Your task\\
    Sentence by sentence, analyze the mechanism discussed in each sentence. \\
    \# Mechanisms involved\\
    You can only name one mechanism for each sentence as follows. You should only include one of the below:\\
    \#\#\# Start of mechanisms \\
    \# Altruism \\
    Definition: Altruism is the principle and practice of concern for the well-being and/or happiness of other humans or animals above oneself. \\ 
    Example: By choosing B1, A will get more in their payoff. \\ 
    ... \{ Other Mechanisms and Examples\}\\
    \#\#\# End of Mechanisms \\ 
    \# The text you need to analyze \{TEXT\} \\ 
    \# Format\\
    You first think step by step, then put your answer as a list, put the list in <mec></mec>. For example <mec>[`understanding', `theory of mind', `altruism']</mec>.
    }}
    }
}

\clearpage
\newpage
\section{Algorithm for Tree-Based Visualizing of Reasoning Processes}
\label{appendix:tree_based}
We present the algorithm for visualizing the reasoning process in Section~\ref{subsubsec:tree_based}. 
\begin{algorithm}
\caption{Build Decision Tree from LLM Responses}\label{alg:build_tree}
\begin{algorithmic}[1]\scriptsize
\setlength{\baselineskip}{10pt}  

\Require A list of responses indexed by $i$, where each $R_i = (R^{\text{cot}}, R^{\text{act}})$
\Require A codebook $\mathcal{C}$ that maps sentences to values

\State Initialize decision tree $\mathcal{T}$
\State $\mathcal{T}.\text{root} \gets$ New TreeNode

\For {each response $R_i$}
    \State Initialize empty list $V_i = \{\}$

    \Comment{Process reasoning $R^{\text{cot}}$}
    \For {each sentence $x_t$ in $R^{\text{cot}}$}
        \State $v \gets \text{codebook lookup of } x_t \text{ in } \mathcal{C}$
        \If {$v$ is not in $\mathcal{C}$}
            \State $v \gets \text{``understanding''}$
        \EndIf
        \If {$v \neq V_i[-1]$} \Comment{If $v$ is not equal to the last element in $V_i$}
            \State Append $v$ to $V_i$
        \EndIf
    \EndFor

    \Comment{Process action $R^{\text{act}}$}
    \State Extract action type $a \in \{B1, B2\}$ or, $\{C1, C2\}$ from $R^{\text{act}}$
    \State Append $a$ to $V_i$

    \Comment{Update tree $\mathcal{T}$ with $V_i$}
    \State Set current node to $\mathcal{T}.\text{root}$

    \For {each value $v \in V_i$}
        \If {current node has no child with value $v$}
            \State Create new child node for $v$
        \EndIf
        \State Move to the child node corresponding to $v$
        \State Increment the count of this node
        \State Increment the count for the decision  $V_i[-1] \in \{B1, B2\}$ (or $\{C1, C2\}$) of this node
    \EndFor

\EndFor

\end{algorithmic}
\end{algorithm}

\clearpage
\newpage
\section{Heterogeneity and Prompt Sensitivity Analyses}
\label{Appendix:Heterogeneity_robustness}
A key challenge in contemporary research on LLMs is the proper design of prompts. Mathematically, this involves defining a set of prompts, represented as environmental states $\mathbb{S}$, that capture the general responses of LLMs when exhibiting preferences in social interactions. The understanding and value $U$ and $V$ are generated according to $\mathbb{P}_{\theta}(\cdot,\cdot | S)$, and actions $A$ drawn from $\mathbb{P}_{\theta}(\cdot | U,V,S)$, for each state $S$ in $\mathbb{S}$. These are supposed to represent the LLM's general understanding, values, or actions based on the underlying processes described in Equations~\eqref{eq:uv} and~\eqref{eq:SUVA} if $\mathbb{S}$ is properly designed.

In this appendix, we perform sensitivity analyses to evaluate how variations in prompt design affect our findings. Due to space constraints, we focus on the most capable models from each model family: GPT-4, LLaMA (3-70B), and Mixtral (8x7B).

\subsection{Sensitivity to Incentive Structure}
\label{appendix:incentive}
Understanding incentive structure in LLM experiments is crucial because, unlike human lab experiments on platforms like Prolific or Mechanical Turk, where incentive structures are based on well-defined opportunity costs like time and money, LLMs do not share the same motivations. Human participants understand the value of their compensation, which helps standardize their responses. In contrast, LLMs do not have personal motivations or an understanding of compensation, making it essential to carefully design and test incentives that align with their operational principles, such as task optimization and performance measures. Without this, the reliability and robustness of LLM responses may be compromised, as they do not react to incentives in the same way humans do.

To examine the effect of incentive structure, we vary two factors in the system prompt: income levels and incentive conversion rates. This allows us to explore how these factors might influence decision-making in LLMs. Income levels correspond to the 10th, 50th, and 90th percentiles in the 2023 US labor market,\footnote{\url{https://dqydj.com/household-income-percentile-calculator/}} with annual incomes of \$17,000, \$74,202, and \$216,056, respectively. The incentive conversion rate determines how many dollars are awarded per 100 points earned in the experiment, with conversion values of \$0.01, \$1, \$100, and \$10,000. Examining these factors together is crucial because income levels might affect how LLMs perceive the value of these incentives. In human subject experiments, for instance, lower-income participants might be more sensitive to changes in monetary rewards, potentially altering their behavior in ways that would differ from higher-income participants~\citep{benabou2006incentives}; it is unclear whether similar phenomena would be observed from LLMs.

\subsubsection{Points-to-monetary-reward conversion rates} We first analyze the robustness of the incentive in terms of the points-to-monetary-reward conversion rates. 
In the context of group identity, the impact of wage variation on distributional preferences shows minimal changes. Across all wage levels (17,000 to 216,056), group identity consistently influences the models’ social welfare preferences. Models such as GPT-4 and LLaMA 3-70b demonstrate a continued preference for social welfare over self-interest when interacting with ingroup members, regardless of wage.

However, the presence of group identity does not amplify or mitigate competitive or difference aversion tendencies significantly across wage levels (Figure~\ref{fig:incentive_group_conv}). Group identity, while maintaining stable effects, does not interact dynamically with wage structures in ways that alter reciprocity or preference shifts, reinforcing that group membership is a robust determinant of behavior independent of wage-based incentives.

For direct and indirect reciprocity (Figure~\ref{fig:incentive_reciprocity}), group identity remains unaffected by changes in wage. Whether the opponent helps or misbehaves, the cooperative behaviors directed toward ingroup members remain stable. Ingroup matches continue to receive similar levels of cooperation across all wage variations, suggesting that group identity maintains its influence even when wage incentives are adjusted.

\subsubsection{Income level relative to incentive structure}

When considering conversion rates (from 0.01 to 10,000), group identity again proves stable in its effects across all examined social preferences. As shown in Figure~\ref{fig:incentive_group_wage}, models like GPT-4 and LLaMA 3-70b display consistent social welfare preferences for ingroup members, showing that shared group identity maintains its significance across conversion levels. Conversion does not induce any significant changes in competitive behavior, difference aversion, or self-interest for ingroup participants, further underscoring the robustness of group identity.
In the reciprocity scenarios (Figure~\ref{fig:incentive_reciprocity}), whether the opponent helped or misbehaved, cooperative behavior toward ingroup members remains unaffected by conversion rates. Group identity continues to play a primary role in guiding social responses, indicating that conversion rates do not weaken or strengthen the underlying group dynamics.

\begin{figure}
    \centering
    \subfloat[Incentive structure on group identity by varying wage \label{fig:incentive_group_wage}]{\includegraphics[width=1\linewidth]{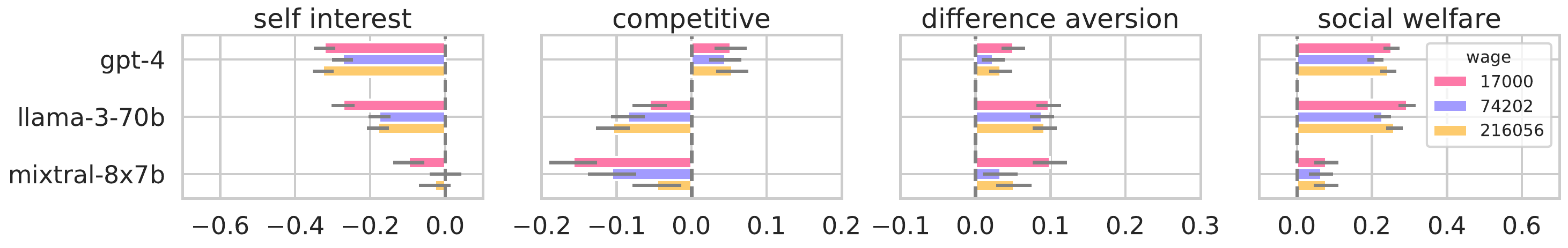}}\\
    \subfloat[Incentive structure on group identity by varying conversion \label{fig:incentive_group_conv}]{\includegraphics[width=1\linewidth]{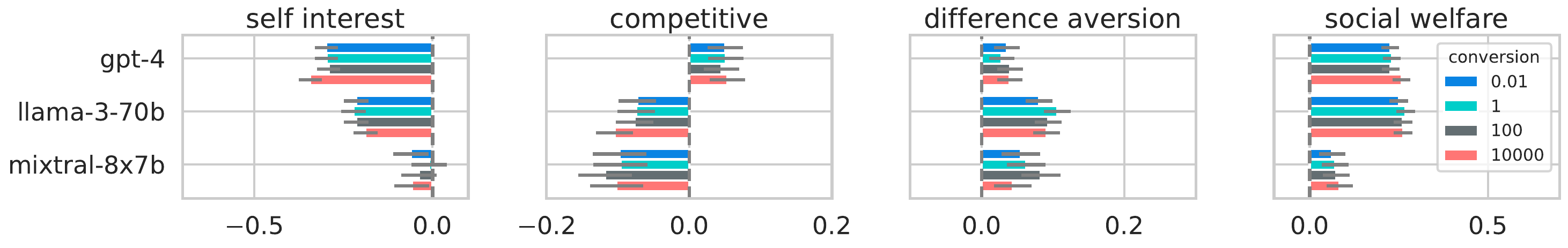}}\\
    \subfloat[Incentive structure on direct and indirect reciprocity \label{fig:incentive_reciprocity}]{\includegraphics[width=1\linewidth]{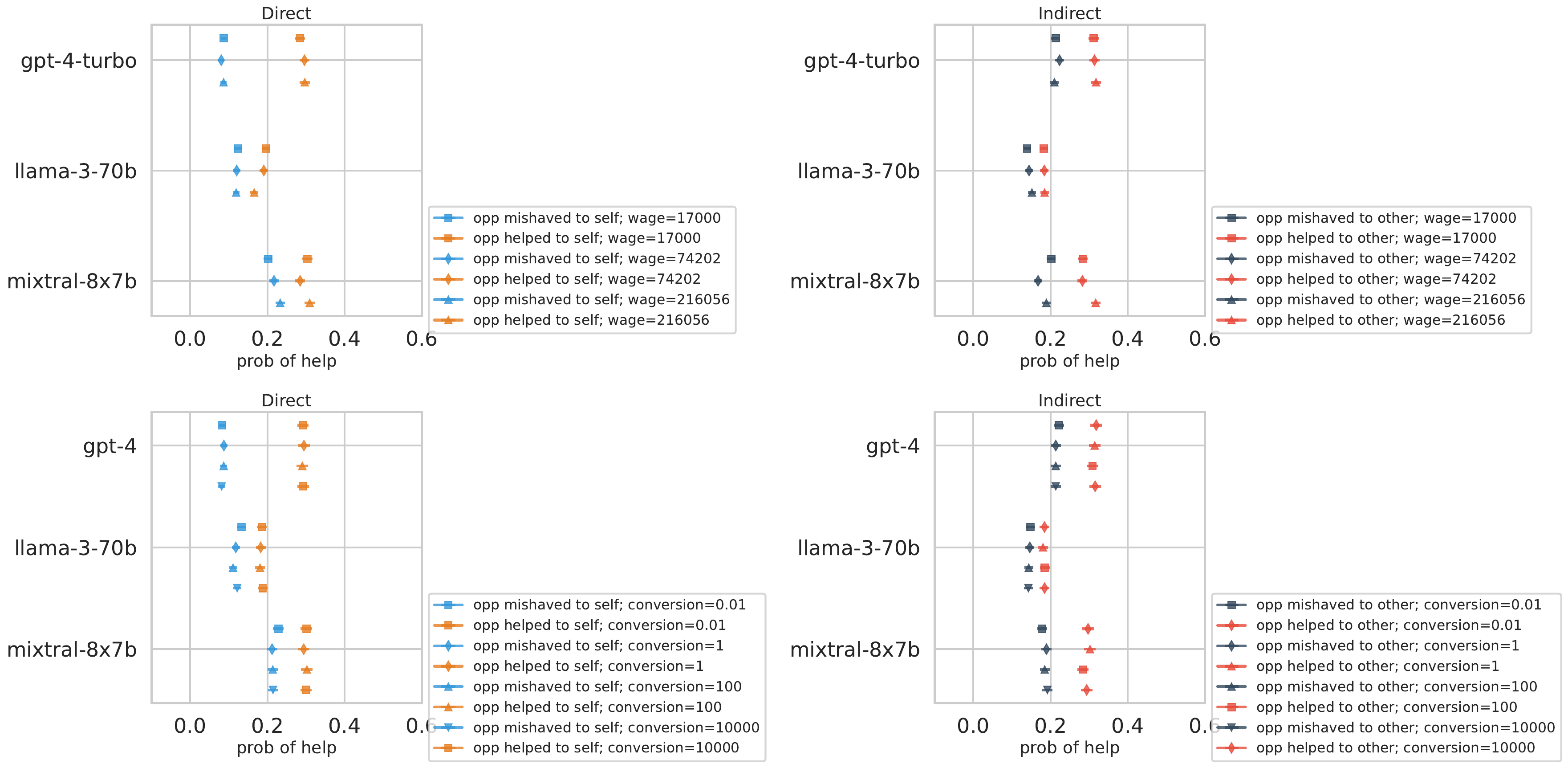}}
    \caption{Effect of incentive structures (wage and conversion) on LLMs' group identity and reciprocity}
    \label{fig:appendix_incentive}
\end{figure}

\subsection{Persona}
\label{appendix:persona}
Persona variables, including demographic and behavioral factors, are crucial in shaping LLMs' responses across diverse contexts. These variables can be easily adjusted to tailor the model's output for specific purposes. However, the literature on persona development and application in LLMs faces several challenges, such as how personas should be defined and implemented for specific purposes.

In our context, we design the persona based on the dataset in~\cite{chen2009group}. This paper provides detailed demographic information, including gender, grade, and major of the student participants, as well as behavioral data, such as whether the students have donation or volunteering experience and the strategies they employed in scenarios like dictator games.

We design each persona by incorporating basic demographic information (such as age, gender, and major) and behavioral data, such as whether the participant has engaged in charitable behavior and the strategies they employed. The distribution of these personas follows the sample characteristics in~\cite{chen2009group}. For each participant in the study, a post-experiment survey was completed, which included basic information like age and gender. Participants also answered a multiple-choice question regarding the strategy they employed, with options such as ``try to earn as much money as possible for myself" (selfish persona), ``try to earn more money than my match" (comparison persona), ``try to choose equal payoff" (fairness persona), and ``try to earn as much money as possible for me and my match" (welfare persona). In our experiment, for each prompt, we randomly select a participant from the sample and use their persona to craft a system prompt. This encourages the LLM to mimic the behavior and preferences of the selected participant.

We analyze the varying results produced by LLMs with different personas. We find that categories such as age, gender, and donation history do not produce significant variability in outcomes, suggesting these factors may not strongly influence LLM behavior in this context. However, strategic personas, which directly relate to decision-making strategies, significantly influence LLM behavior. These personas prompt the LLMs to adopt the corresponding distributional preferences more strongly.

In Figure~\ref{fig:persona},
we present the heterogeneity of LLM agents with different strategy personas across four dimensions of distributional preference. We observe that, across all three models, LLM agents with a strategy persona show a stronger tendency towards the corresponding distributional preference. For instance, agents with a selfish persona demonstrate a heightened tendency towards self-interest across all models, showing the largest magnitude in this preference. Similar patterns are observed for the other personas, with the largest magnitude typically aligning with the expected distributional preference. The width of the confidence intervals largely depends on the number of LLM agents assigned to each persona, with more common personas like the selfish persona resulting in narrower CIs.

These findings highlight the crucial role of prompt design in shaping LLM behavior. Utilizing strategic personas can effectively guide models to produce responses that align with specific expectations.

\subsection{Additional Ananlyses}
\label{appendix:robustness}

\paragraph{Temperature.}

In the main analysis, we used a temperature setting of 0.2, which constrains the model's variation in text generation, leading to more deterministic outputs. 
To assess the robustness of our findings with respect to temperature, we explore the effects of increasing the temperature to 0.8, which allows for greater variability in the generated text.

The results in Figure~\ref{fig:robustness} indicate that the distributional preferences across the different models—GPT-4, LLaMA (3-70B), and Mixtral (8x7B)—remain largely robust to this change in temperature. Compared to the results at the lower temperature setting, there are no statistically significant differences in self-interest, competition, or social welfare metrics across the models, suggesting that the increase in temperature does not significantly impact their decision-making processes. The only minor exception is observed in the LLaMA (3-70B) model, where difference aversion slightly increases with the higher temperature, though the difference is marginal. Overall, these findings suggest that temperature adjustments have minimal effects on the behavior of these models, with their distributional preferences remaining stable across different temperature settings.

We also examine effects of temperature on group identity and reciprocity.
When temperature is varied between 0.2 and 0.8, the results for group identity and reciprocity are mostly consistent with the main findings in the distributional preferences. For both GPT-4 and LLaMA (3-70b), there are no significant differences between the two temperature settings. In both cases, the models maintain stable behavior in terms of social welfare, competition, and self-interest when interacting with ingroup members.

However, for Mixtral (8x7B), there are slight differences in behavior between the two temperature settings. Under higher temperature (0.8), there is a small increase in competition and a slight decrease in social welfare orientation. Despite these minor changes, the overall qualitative patterns remain the same, suggesting that temperature does not substantially affect the core group identity effects.
For reciprocity, there is a minor variation, with slightly more responsive behavior (higher probability of help) observed at temperature 0.2. However, this difference is small, and overall, the qualitative behavior remains the same, highlighting that temperature has minimal impact on reciprocity patterns across all models.

\paragraph{Importance of Chain-of-Thought Prompting.}

We next examine the impact of removing chain-of-thought (CoT) prompting, specifically the instruction to ``\texttt{Please reason step-by-step.}" We assess the results under this modified prompt structure to evaluate how the absence of explicit reasoning prompts influences the model's outputs. Adding CoT prompting increases the alignment between LLMs and human, and enhances the correctness of outputs~\citep{goli2023language, nayab2024concise}. However, CoT reasoning often results in longer outputs, which can increase the time and computational cost required for generating responses. 

The results of the CoT analysis in Figure~\ref{fig:robustness} reveal distinct impacts on the distributional preferences of the models, with each responding differently to the introduction of CoT reasoning. For GPT-4, CoT leads to an increase in self-interest and a decrease in social welfare, indicating a shift towards prioritizing individual outcomes at the expense of broader societal considerations. In LLaMA (3-70B), CoT results in heightened competition and difference aversion, along with a decrease in social welfare, suggesting a move towards more competitive and self-focused behavior that may undermine collective benefits. In contrast, Mixtral (8x7B) shows reduced variance without CoT, alongside increased competition and difference aversion, coupled with a decline in social welfare. This pattern suggests that while Mixtral (8x7B) becomes more competitive and equitable, it also tends to sacrifice social welfare outcomes.

When comparing results with and without CoT reasoning, there are more noticeable differences, especially for GPT-4 and LLaMA 3-70b. Specifically:
\begin{itemize}
    \item Regarding group identity, GPT-4 shows more favor towards ingroup matches when CoT is disabled. In terms of reciprocity, GPT-4 exhibits less reciprocity behavior when CoT is disabled.

    \item LLaMA (3-70B) shows the opposite trend for group identity: it demonstrates fewer prosocial actions when CoT is disabled. Regarding reciprocity, LLaMA (3-70B) exhibits stable levels of direct reciprocity but shows more indirect reciprocity when CoT is disabled.

    \item For Mixtral (8x7B), there are fewer prosocial actions and a greater demonstration of both direct and indirect reciprocity when CoT is disabled.

\end{itemize}

\end{APPENDICES}

\end{document}